\pdfoutput=1
\documentclass[11pt]{article}

\usepackage[final]{acl}

\usepackage{tabularx}
\usepackage{multirow}
\usepackage{booktabs}
\usepackage{graphicx}
\usepackage{colortbl}
\usepackage{array}
\usepackage{longtable}

\usepackage{times}
\usepackage{amsmath}
\usepackage{tcolorbox}
\usepackage{verbatim}
\usepackage{latexsym}
\usepackage{algorithm2e}
\usepackage{xcolor}
\usepackage{color}
\usepackage{makecell}

\usepackage{algpseudocode}
\newcommand{\method}{\textsc{TabDSR}\xspace}
\newcommand{\benchmark}{\textsc{CalTab151}\space}
\definecolor{mygreen}{rgb}{0.1, 0.6, 0.3} % 自定义绿色
\definecolor{myred}{rgb}{0.8, 0.1, 0.2}  % 自定义红色
\definecolor{shade}{rgb}{0.92,0.92,0.92}

\usepackage[T1]{fontenc}
\usepackage[utf8]{inputenc}

\usepackage{microtype}

\usepackage{inconsolata}

\def\colornum[#1][#2][#3]#4{
  \newcommand{\Min}{#1} % 最小值
  \newcommand{\Max}{#2} % 最大值
  \newcommand{\Mid}{#3} % 中间值(可选）
  \pgfmathsetmacro{\Percent}{max(min(100.0*(#4 - \Min)/(\Max-\Min),100.0),0.00)}
  \hspace{-0.3em}\colorbox{yellow!\Percent!white}{#4}
}

\definecolor{lightyellow}{rgb}{1.0, 1.0, 0.7}
\definecolor{mediumyellow}{rgb}{1.0, 0.9, 0.3}
\definecolor{darkyellow}{rgb}{1.0, 0.8, 0.1}

\title{\method: Decompose, Sanitize, and Reason for Complex Numerical Reasoning in Tabular Data}

\author{
 \textbf{Changjiang Jiang\textsuperscript{1}},
 \textbf{Fengchang Yu\textsuperscript{1}},
 \textbf{Haihua Chen\textsuperscript{2}},
 \textbf{Wei Lu\textsuperscript{1}},
 \textbf{Jin Zeng\textsuperscript{1,3}}
 \\
 \textsuperscript{1}Wuhan University,\\
 \textsuperscript{2}University of North Texas,\\
 \textsuperscript{3}Hubei University of Economics
 \\
 \small{
   \textbf{Correspondence:} \href{mailto:yufc2002@whu.edu.cn}{yufc2002@whu.edu.cn}
 }
}

\begin{document}
\maketitle
\begin{abstract}

Complex reasoning over tabular data is crucial in real-world data analysis, yet large language models (LLMs) often underperform due to complex queries, noisy data, and limited numerical capabilities. To address these issues, we propose \method, a framework consisting of: (1) a query decomposer that breaks down complex questions, (2) a table sanitizer that cleans and filters noisy tables, and (3) a program-of-thoughts (PoT)-based reasoner that generates executable code to derive the final answer from the sanitized table. To ensure unbiased evaluation and mitigate data leakage, we introduce a new dataset, CalTab151, specifically designed for complex numerical reasoning over tables. Experimental results demonstrate that \method consistently outperforms existing methods, achieving state-of-the-art (SOTA) performance with 8.79\%, 6.08\%, and 19.87\% accuracy improvement on TAT-QA, TableBench, and \method, respectively. Moreover, our framework integrates seamlessly with mainstream LLMs, providing a robust solution for complex tabular numerical reasoning. These findings highlight the effectiveness of our framework in enhancing LLM performance for complex tabular numerical reasoning. Data and code are available upon request.
\end{abstract}

\section{Introduction}

\begin{figure}[t]
  \centering
  \includegraphics[width=\columnwidth]{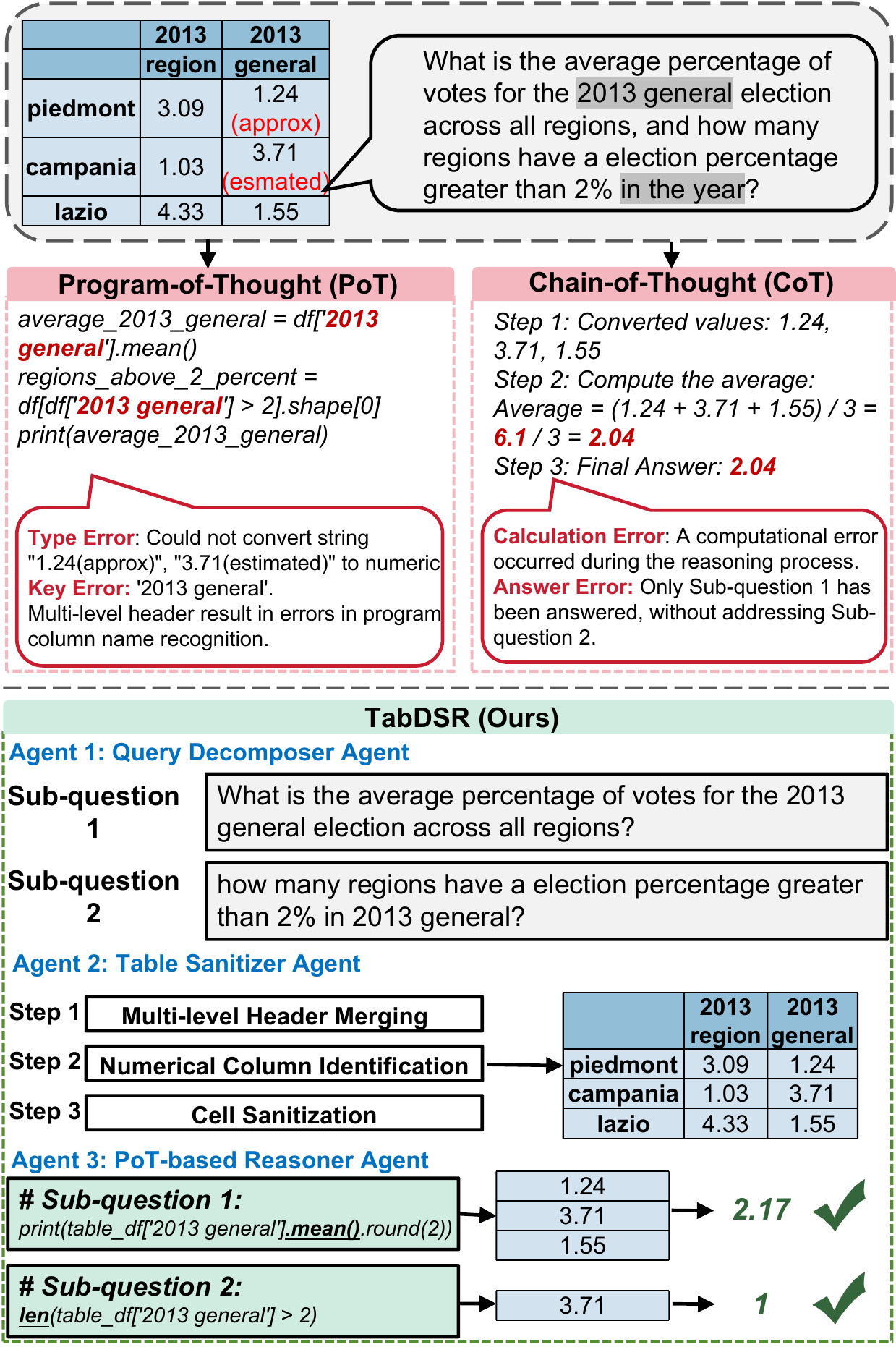}
  \caption{Illustration of the comparison between PoT, CoT, and the proposed \method.}
  \label{fig:problem}
\end{figure}

Table Question Answering (TQA) requires extracting and reasoning over numerical information from tabular data to produce correct answers. It has found broad application in financial TQA~\citep{chen2021finqa,zhu-etal-2021-tat} and mathematical reasoning with tabular data~\citep{lu2023dynamic}. Although large language models (LLMs) exhibit strong general reasoning capabilities, real-world TQA remains challenging. TQA questions often demand multi-hop reasoning~\citep{biran-etal-2024-hopping}, involving multiple calculation steps, and visual table conversions can introduce noise, further degrading performance. Recently, agent-based approaches, such as Deep Research~\citep{zhang2025agentorchestrahierarchicalmultiagentframework} and GUI Agents~\citep{yuan2025enhancing}, have been explored to handle multi-step reasoning and complex multi-hop tasks.

Approaches to TQA numerical reasoning generally fall into three categories: pre-trained models, fine-tuning LLMs, and prompt-based LLMs. Pre-trained and fine-tuned models typically require large amounts of high-quality TQA data, and they struggle to generalize to new or unseen tasks. As a result, prompt-based methods have become the mainstream solution. Nonetheless, as illustrated in Figure~\ref{fig:problem}, they still face three critical challenges: (1) \textbf{Multi-hop Complexity}: Even LLMs with strong reasoning abilities can fail to answer multi-hop questions accurately. (2) \textbf{Data Quality and Structure}: Program-of-Thought (PoT) approaches~\citep{chen2023program} rely on clean and consistently typed table columns; mixed-type columns (e.g., ``1.24(approx)'' and ``1.55'' in the same column) can trigger runtime errors. (3) \textbf{Limited Numerical Computation}: Current LLMs do not truly compute but rather mimic numerical procedures seen in training data~\citep{mirzadeh2024gsm}.

Humans, by contrast, excel at table reasoning via a three-step process: (1) decompose complex questions into simpler sub-questions, (2) interpret the table’s structure and semantics, and (3) extract relevant data and perform precise calculations. Inspired by this process, we introduce \method, a prompt-based framework tailored for numerical reasoning with complex tables. As depicted in Figure~\ref{fig:problem}, \method employs three agents that mirror human reasoning: (1) A \textbf{Query Decomposer Agent} that splits the original query into tractable sub-questions, (2) A \textbf{Table Sanitizer Agent} that refines the table to ensure a clean, machine-readable format, (3) A \textbf{PoT-based Reasoner Agent} that generates and executes programs to derive the final answer.

Evaluating TQA methods is further complicated by data leakage in existing datasets~\citep{deng-etal-2024-investigating}, which can obscure the true capabilities of LLMs. To address this, we propose \benchmark, a new dataset specifically designed to assess complex numerical reasoning in TQA while minimizing data leakage risks. This provides a more reliable benchmark for evaluating prompt-based frameworks, ensuring a fair comparison of different LLM-based solutions for table-based numerical reasoning.

Our contributions are summarized below:

\begin{itemize}
    \item We propose \method, a framework that combines Query Decompose Agent, Table Sanitizer Agent and PoT-based Reasoner Agent, specifically designed to improve the performance of LLMs in numerical reasoning tasks within TQA.
    \item We conducted a comprehensive analysis of our approach in several open-source and closed-source LLMs. \method significantly improves the performance of LLMs in numerical reasoning within TQA tasks. Experimental results across different LLMs demonstrate the transferability of our method.
    \item We construct \benchmark, a novel dataset in complex numerical reasoning TQA task, to avoid potential data leakage that could undermine the reliability of the reported metrics on the public datasets.
\end{itemize}

\begin{figure*}[t]
  \centering
  \includegraphics[width=\textwidth]{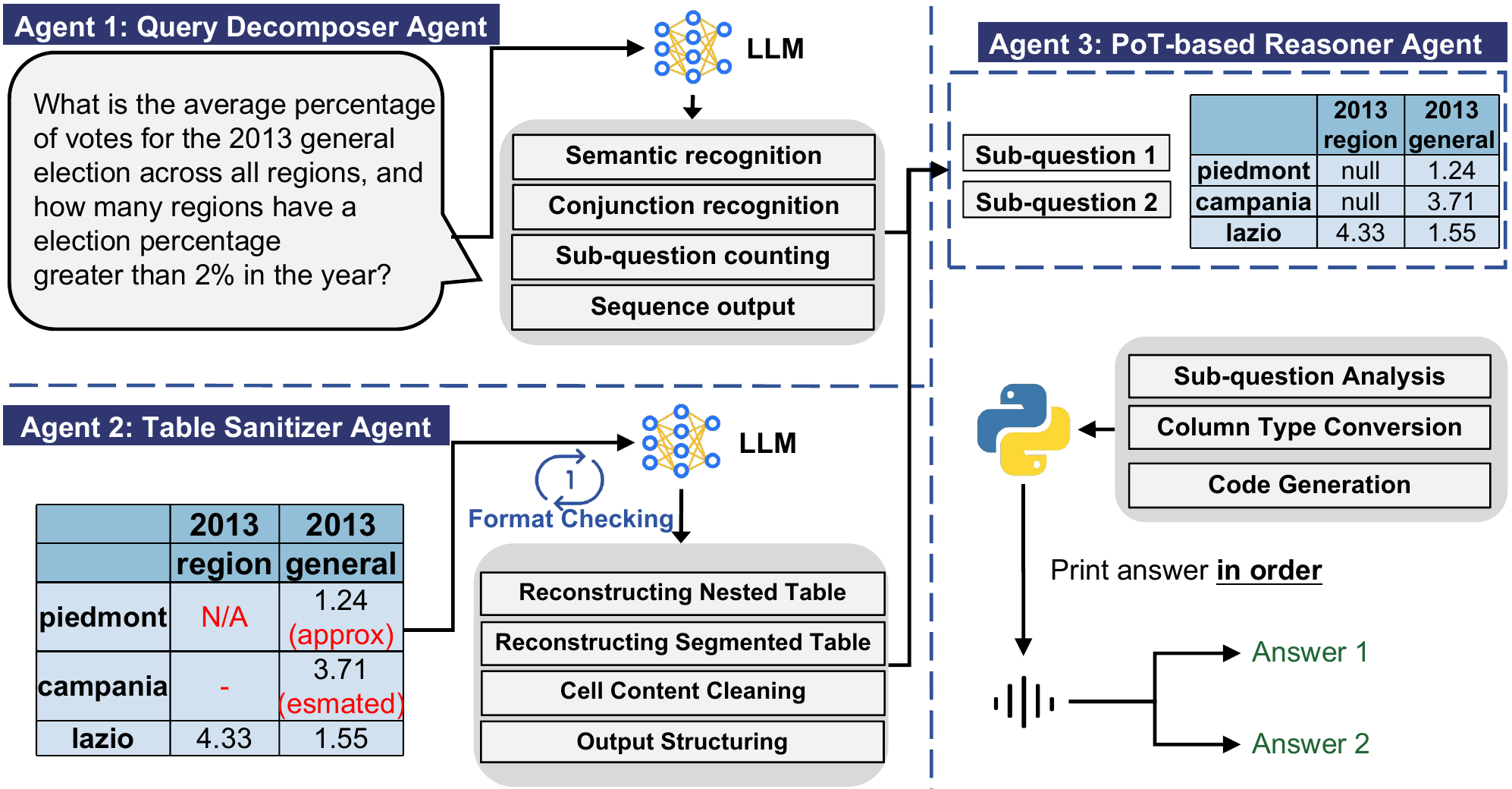}
  \caption{\method's collaborative pipeline: Synchronous execution of query decomposer agent and table sanitizer agent for complex numerical reasoning via PoT.}
  \label{fig:TabDSR}
\end{figure*}

\section{Related Work}

% \subsection{Pre-trained models}

\textbf{Pre-trained models}. Early work on Table Question Answering (TQA) often relied on large pre-trained models fine-tuned for specific downstream tasks. Given a table and a question, some methods generate executable SQL queries to extract the answer~\citep{liu2022tapex,jiang-etal-2022-omnitab}, while others directly predict the answer text~\citep{herzig-etal-2020-tapas}. Although these approaches have shown promising results, they typically require large-scale, high-quality tabular datasets and significant computational resources for training. As a result, they face challenges in domain adaptation and often struggle to generalize to new or unseen tables.

% \subsection{Fine-tuning LLMs}

\textbf{Fine-tuning LLMs}. Recent studies have shown that directly fine-tuning large language models (LLMs) can yield strong performance on table reasoning tasks~\citep{badaro-etal-2023-transformers}, as seen in systems like TableLlama~\citep{zhang-etal-2024-tablellama}, TableGPT2~\citep{10.1145/3654979}, and TableLLM~\citep{zhang2024tablellmenablingtabulardata}. These fine-tuned models obviate the need for elaborate prompt design by framing the task as a single-round QA process. However, this approach compresses the reasoning steps into a single inference pass, limiting traceability for numerical calculations.

% \subsection{Prompt-based LLMs}
% \label{sec:relate_work_prompt}

\textbf{Prompt-based LLMs}. For TQA with LLMs, direct prompting (DP) feeds the table and question into an LLM for a single-step answer without intermediate reasoning. In contrast, \textbf{Chain-of-Thought (CoT) prompting} compels LLMs to generate step-by-step reasoning before producing the final answer. Within CoT, two main CoT variants are textual chain-of-thought (TCoT) and symbolic chain-of-thought (SCoT). TCoT appends prompts like ``Let's think step by step!'' \citep{kojima2022large}, and SCoT uses symbolic commands to iteratively refine results \citep{wu2024tablebench}. An emerging trend in TQA is to merge CoT prompting with multiple LLM calls, as demonstrated by MFORT-QA\citep{guan2024mfort}, which first employs few-shot prompts to retrieve relevant tables, then applies CoT prompts to guide the reasoning process. However, GSM-Symbolic\citep{mirzadeh2024gsm} reveals limitations in LLM numerical reasoning: systematically altering numerical values in math problems degrades model performance. This finding highlights that CoT prompting alone cannot ensure accurate numerical calculations.

Another popular approach is \textbf{Program-of-Thought (PoT)} prompting. PoT-based methods employ LLMs to generate executable code, addressing the models’ inherent computational limitations. However, as illustrated in Figure~\ref{fig:problem}, these methods rely on clear questions, correct table structures, and consistent column types. Parallel lines of research on question decomposition often train sequence-to-sequence models to split a complex question into smaller sub-questions~\citep{perez2020unsupervised,zhang-etal-2019-complex}, but this requires extensive manual labeling. In TQA, question decomposition is frequently combined with table-level preprocessing. For instance, TabSQLify\citep{nahid-rafiei-2024-tabsqlify} condenses large tables to reduce contextual overhead, and MIX-SC\citep{liu-etal-2024-rethinking} integrates a Python interpreter with ReAct~\citep{yao2023react} iterative reasoning to refine its outputs. Similarly, Chain-of-Table~\citep{wang2024chainoftable} dynamically plans operations based on sub-task selection.

Despite these advances, two challenges persist: (1) existing methods often fail to explicitly address multi-hop questions, risking incomplete or incorrect answers, and (2) they overlook column-type inconsistencies, which can cause LLM-generated code to misinterpret text noise as numerical data and trigger computation errors.

\section{\method}

\paragraph{Task Formulation} The goal of Table Question Answering (TQA) is to predict an answer $A$ given two table strings: a \textbf{table} $T$ and a \textbf{question} $Q$. As illustrated in Figure~\ref{fig:TabDSR}, our \method framework addresses TQA’s numerical reasoning challenges via three specialized agents, each responsible for a key component of the reasoning pipeline.

\subsection{Query Decomposer Agent}

Complex numerical reasoning often requires \textbf{multi-hop} interpretation of the question. Single-turn models frequently fail to capture this complexity, resulting in incomplete or inaccurate answers. Previous work underscores the importance of proper question understanding in TQA.

Existing prompt-based methods typically feed both the question and the table into an LLM. For example, DATER~\citep{10.1145/3539618.3591708} uses the prompt ``Decompose questions into sub-questions and transform them into a cloze-style format'' However, these decomposition instructions can be vague, yielding inconsistent sub-question granularity. Moreover, given that tables are often much larger than the question text, crucial details in the question may be overshadowed by the table content during decomposition.

To address these issues, our Query Decomposer Agent takes only the question as input and ignores the table entirely. By significantly reducing the prompt length, we ensure that pertinent details in the question are more likely to be retained. Concretely, we design the prompt to decompose the query based on textual cues such as conjunctions (``and'', ``or'') and punctuation (commas), treating each segment as an independent sub-question.

We further mitigate potential LLM hallucinations by drawing inspiration from Chain-of-Thought (CoT) reasoning~\citep{10.5555/3600270.3602070}. Specifically, we instruct the model to output the number of sub-questions in a list format, accompanied by a carefully chosen example from a complex TQA query. This example demonstrates the desired level of decomposition and guides the model’s reasoning process.

\begin{figure}[t]
  \centering
  \includegraphics[width=\columnwidth]{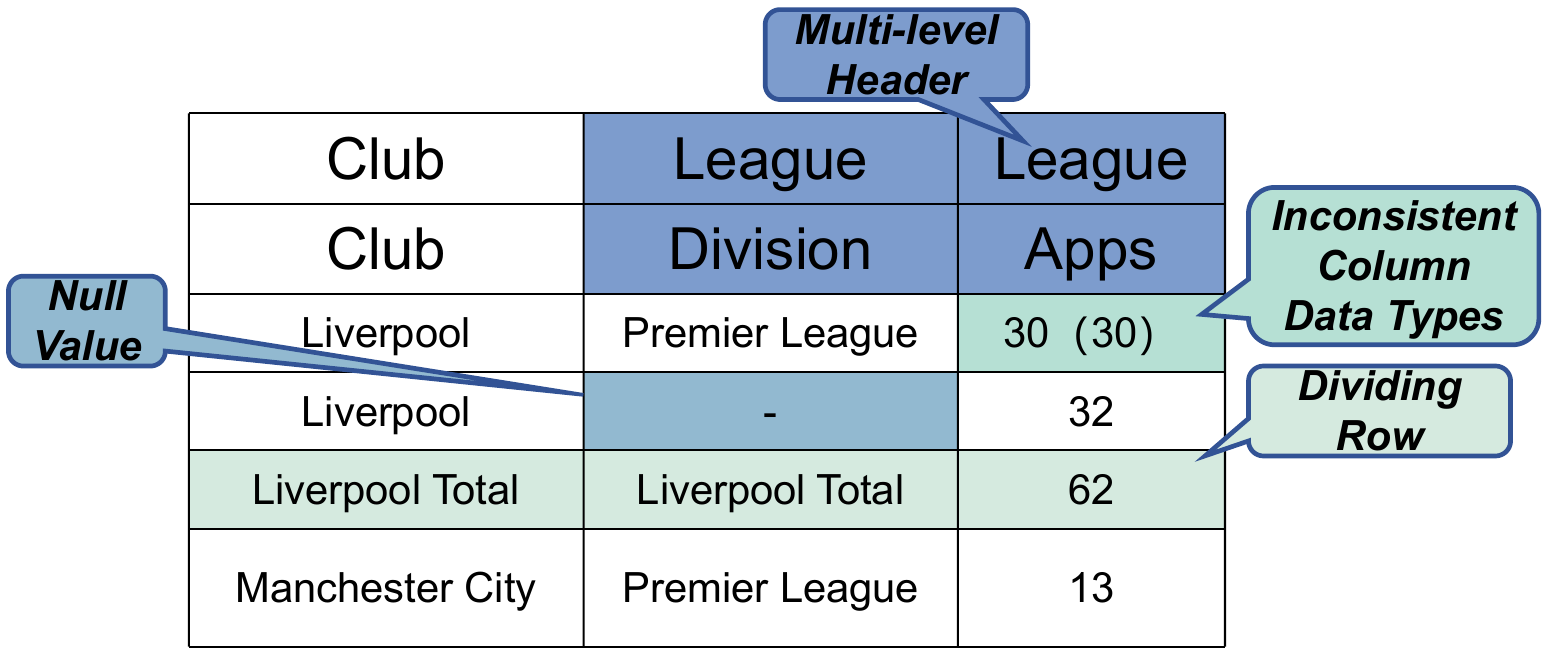}
  \caption{Example of a noisy tabular data; The original table is sourced from the TableBench~\citep{wu2024tablebench}; The table has multi-level headers, which seem to introduce errors and noise due to visual table conversions; For the sake of presentation, we manually removed certain rows and columns and modified a few cells.}
  \label{fig:tableerror}
\end{figure}

\subsection{Table Sanitizer Agent}

Beyond the complexity of the question text, TQA tasks are often complicated by textual tables that lose the visual cues crucial for understanding hierarchical or segmented data. These tables can be lengthy, contain redundant rows and columns, include null values, or introduce noise through the conversion process from a visual to a text-based format. To tackle these issues, our \textbf{Table Sanitizer Agent} optimizes both the structure and content of textual tables.

\textbf{Structural Optimization}, which consists of two scenarios: (1) Reconstructing Nested Tables (Multi-level Headers). In text format, headers from multi-level tables can be split into separate lines, obscuring their logical relationships. We prompt the model to detect these nested headers and merge them based on their semantic similarities. (2) Reconstructing Segmented Tables. Tables sometimes appear in sections separated by blank rows or dividing lines, and these visual cues are lost in plain text. We enhance the prompt to identify these segmentation rows and either remove or extract them, depending on the query requirements.

\textbf{Content Optimization}, mainly focuses on Cell Content Cleaning. Subsequent reasoning steps rely on valid cell entries. Accordingly, we instruct the model to remove extraneous symbols (e.g., ``\%,'' currency symbols, commas), explanatory notes, emojis, and other non-numeric characters. We then convert numerical data into integer or float formats and standardize blank cells (e.g., ``–'', ``N/A'') to a consistent ``null'' label.

Although some studies simplify tables to reduce complexity~\citep{10.1145/3539618.3591708,nahid-rafiei-2024-tabsqlify}, modern LLMs can effectively handle long inputs. For instance, Qwen2.5:7b supports a 128K context window~\citep{yang2024qwen2}, making the full table content manageable without sacrificing critical information. Consequently, our Table Sanitizer Agent retains the complete table to preserve as much detail as possible for downstream TQA tasks. 

To mitigate potential hallucinations from the LLM that could lead to table-cleaning errors, we incorporate a reflection mechanism. Specifically, the cleaned table string is first validated by a Python parser. If the parser fails to read the table, the resulting error message, along with the newly generated string, is fed back into the prompt as contextual guidance for the LLM to regenerate a corrected version. To prevent infinite loops, we limit this regeneration process to a single additional iteration in our current experiments. The overall workflow is illustrated in Figure~\ref{fig:TabDSR}.

\subsection{PoT-based Reasoner Agent}

Having decomposed the original question into sub-questions and sanitized the table, we now have well-defined queries and clean data ready for computation. However, studies such as \citet{mirzadeh2024gsm} indicate that LLMs alone often struggle with precise numerical reasoning. To circumvent this limitation, our PoT-based Reasoner Agent employs Program-of-Thought (PoT) techniques to generate Python code that performs the necessary calculations on the sanitized table.

While some approaches~\citep{wang2024chainoftable,nahid-rafiei-2024-tabsqlify} rely on SQL for table computations, our method prioritizes simplicity and computational efficiency. Specifically, we load the sanitized data into a Pandas DataFrame and leverage its flexible APIs to handle tasks such as filtering, aggregation, and arithmetic operations. To ensure robustness, we (1) extract relevant data into the DataFrame based on each sub-question, (2) generate Python code that executes calculations specific to these sub-questions, (3) restrict certain Pandas methods to avoid inconsistencies across versions, (4) validate data formats through consistency checks, minimizing the risk of errors caused by unexpected input types.

After computing the results for each sub-question, we reassemble them in a logical sequence—considering any dependencies between questions—to produce the final TQA answer. This modular pipeline ensures that the correctness of PoT-based reasoning is maximized by the clarity of the sub-questions and the cleanliness of the tabular data, underscoring the importance of the first two agents in our framework.

\begin{table*}[t]
\centering
% \footnotesize
% \vspace{-3mm}
\resizebox{\textwidth}{!}{%
\begin{tabular}{lcccccc}
\toprule
\multirow{2}[2]{*}{\textbf{Methods}} & \multicolumn{2}{c}{\textbf{TAT-QA}} & \multicolumn{2}{c}{\textbf{TableBench}} &
\multicolumn{2}{c}{\textbf{\benchmark}}
\\
\cmidrule(lr){2-3} \cmidrule(lr){4-5} \cmidrule(lr){6-7} &
\textbf{Acc} & \textbf{ROUGE-L} & \textbf{Acc} & \textbf{ROUGE-L} & \textbf{Acc} & \textbf{ROUGE-L} \\
\midrule
\textit{Pretrained-models} \\
\quad TAPEX~\citep{liu2022tapex} & 12.56 & 17.23 & 24.10 & 27.98 & 8.61 & 12.38  \\
\quad OmniTab~\citep{jiang-etal-2022-omnitab} & 13.61 & 17.32 & 27.74  & 31.73 & 9.93 & 14.28  \\
\midrule
\textit{Fine-tuning LLMs} \\
\quad TableGPT2-7B~\citep{su2024tablegpt2largemultimodalmodel} & 9.58  & 9.71  &  44.57  &  45.95 & 14.24 & 15.11 \\
\quad TableLLM-13B@PoT~\citep{zhang2024tablellmenablingtabulardata} & 3.67 & 3.85 & 40.16 & 41.95 & 8.61 & 10.07 \\
\quad TableLLM-13B@DP~\citep{zhang2024tablellmenablingtabulardata} & 35.56 & 37.44 & 42.91 & 44.21 & 24.17 & 28.32 \\
\quad TableLlama-7B~\citep{zhang-etal-2024-tablellama} & 27.84 & 33.51 & 23.70 & 26.75 & 7.28 & 11.50 \\
\midrule
\textit{Prompt-based LLMs} \\
\quad Chain-of-Table~\citep{wang2024chainoftable} & 34.89 & 41.33  &  35.80  &  37.87 & \underline{25.83} & \underline{31.60} \\
\quad TabSQLify\textsubscript{col+row}~\citep{nahid-rafiei-2024-tabsqlify} & \underline{51.62} & \underline{58.17} & 43.04 & 46.97 & 23.51 & 29.62 \\
\quad MIX-SC~\citep{liu-etal-2024-rethinking} & 30.77 & 34.06 & \underline{46.67} & \underline{49.41} & 16.56 & 20.78 \\
\quad $E^{5}$\textsubscript{code}~\citep{zhang-etal-2024-e5} & 40.1 & 44.65 & 9.4 & 11.09 & 8.94 & 10.42 \\
\quad $E^{5}$\textsubscript{zero-shot}~\citep{zhang-etal-2024-e5} & 44.58 & 48.67 & 8.72 & 9.5 & 7.28 & 8.96 \\
\quad NormTab~\citep{nahid-rafiei-2024-normtab} & 21.26 & 22.68 & 37.14 & 38.73 & 10.93 & 12.53 \\
\quad ReAcTable~\citep{DBLP:journals/corr/abs-2310-00815} & 28.72 & 31.27 & 25.27 & 26.31 & 11.59 & 13.38 \\
\midrule
\quad \method (Our Method) & 
  \makecell{\textbf{60.41} \\ \textcolor{myred}{(+8.79)}} & 
  \makecell{\textbf{62.76} \\ \textcolor{myred}{(+4.59)}} &
  \makecell{\textbf{52.75} \\ \textcolor{myred}{(+6.08)}} & 
  \makecell{\textbf{55.39} \\ \textcolor{myred}{(+5.98)}} & 
  \makecell{\textbf{45.70} \\ \textcolor{myred}{(+19.87)}} & 
  \makecell{\textbf{50.57} \\ \textcolor{myred}{(+18.97)}} \\
\bottomrule
\end{tabular}%
}
\caption{Table reasoning results on TAT-QA, TableBench and \benchmark; \textbf{Bold} indicates the best performance; \underline{Underline} indicates the second-best performance; \textcolor{myred}{Red} indicates the improvement measured against the second-best performance method; TableLLM-13B@PoT refers to conducting code execution prompt to generate the final answer; TableLLM-13B@DP refers to conducting direct text answer generation prompt to generate the final answer.}
\label{tab:ResultsTables}%
\end{table*}%

\section{Construction of \benchmark}
\label{sec:caltab151}

To ensure a fair evaluation that mitigates data leakage from existing public datasets, we propose an annotation framework combining LLM-generated queries with human-verified answers (details in Appendix~\ref{appendix:caltab151}). This process culminates in a high-quality numerical reasoning dataset, \benchmark, composed of 151 table samples drawn from TableBench\citep{wu2024tablebench} (84), FinQA\citep{chen2021finqa} (27), TAT-QA\citep{zhu-etal-2021-tat} (32), and AitQa\citep{katsis-etal-2022-ait} (8).

We construct \benchmark\ through the \textbf{six} steps: (1) \textbf{Numerical Perturbation}: To maintain realistic yet varied numeric values, we randomly perturb cell values by $\pm3\%$--$5\%$ of their original values. The prompt can be seen in Figure~\ref{fig:NumericalPerturabation}. (2) \textbf{Cell Noise Addition}:  To simulate natural table noise, we inject context-appropriate symbols (e.g., \$, €, or \%) into randomly chosen numeric columns, aligned with their semantic relevance. (3) \textbf{Structural Randomization}: We enhance structural diversity by shuffling rows or columns and randomly deleting a subset of them, thereby introducing a broader range of table configurations. (4) \textbf{Random Null Value Filling}: To model incomplete data, we replace 2–4 cells with labels such as ``None'', ``Null'', ``N/A'', ``???'', or ``-''. (5) \textbf{Multi-hop Question Generation}: To increase question complexity, we construct an agent for generating multi-hop questions, where each question consists of multiple sub-questions, with each subsequent question depending on the answer to the previous one. The agent guides the model to generate two sub-questions and then merge them into a coherent two-hop question based on natural semantics, ensuring the coherent question is contextually relevant. The Prompt can be seen in Figure~\ref{fig:QueryGeneration}. (6) \textbf{Answer Annotation}: Finally, to ensure data accuracy, human annotators manually calculate and verify answers to all generated multi-hop questions. 

This multi-pronged approach produces a robust and realistic dataset that captures both the structural and semantic challenges of real-world TQA, providing a more reliable benchmark for evaluating LLM-based numerical reasoning.

\section{Experiments and Results}

\begin{table*}[t]
\centering
\resizebox{\textwidth}{!}{%
\begin{tabular}{c|c|cc|cc|cc}
\toprule
 \multirow{2}{*}{\textbf{Model}} & \multirow{2}{*}{\textbf{Method}} & \multicolumn{2}{c}{\textbf{TAT-QA}} & \multicolumn{2}{c}{\textbf{TableBench}} & \multicolumn{2}{c}{\textbf{\benchmark}} \\
 \multirow{-1}{*}{} &  & Acc & ROUGE-L & Acc & ROUGE-L & Acc & ROUGE-L \\
\midrule
% \rowcolor[gray]{.9}
% \multicolumn{7}{c}{\textbf{\textit{Qwen2.5-7B-Instruct}}} \\ 
% \midrule
\multirow{8}{*}{\rotatebox[origin=c]{90}{\textbf{\textit{Qwen2.5-7B}}}} & DP & 29.68 & 35.27 & 16.26 & 19.80 & 8.28 & 12.84 \\
& PoT & 9.34 & 9.70 & 36.02 & 37.57 & 11.59 & 13.15 \\
& TCoT & 55.02 & 62.41 & 29.45 & 34.40 & 19.87 & 27.00 \\
& SCoT & 43.65 & 49.96 & 24.48 & 29.12 & 16.23 & 24.60 \\
& \method\textsubscript{R} & 18.70 & 19.03 & 45.68 & 47.74 & 23.18 & 25.59 \\
& \method\textsubscript{D+R} & 17.69 & 17.91 & 45.14 & 46.86 & 17.88 & 19.30 \\
& \method\textsubscript{S+R} & \textbf{60.50} & \textbf{62.86} & \underline{51.41} & \underline{53.86} & \underline{39.07} & \underline{44.44} \\
& \method\textsubscript{D+S+R} & \underline{60.41} & \underline{62.76} & \textbf{52.75} & \textbf{55.49} & \textbf{45.70} & \textbf{50.57} \\
\midrule
\multirow{8}{*}{\rotatebox[origin=c]{90}{\textbf{\textit{Qwen2.5-Code-7B}}}} & DP & 32.80 & 37.99 & 22.99 & 26.34 & 14.57 & 20.58 \\
& PoT & 20.31 & 20.50 & 40.84 & 42.14 & 5.96 & 6.53 \\
& TCoT & 59.61 & \textbf{66.75} & 42.78 & 46.62 & 22.85 & 31.69 \\
& SCoT & 46.78 & 53.71 & 28.05 & 33.19 & 18.54 & 26.17 \\
& \method\textsubscript{R} & 18.63 & 18.98 & 53.92 & 55.75 & 30.13 & 33.21 \\
& \method\textsubscript{D+R} & 18.09 & 18.45 & 51.37 & 52.71 & 28.15 & 29.51 \\
& \method\textsubscript{S+R} & \textbf{60.96} & \underline{62.56} & \textbf{55.98} & \textbf{58.18} & \underline{46.69} & \underline{50.67} \\
& \method\textsubscript{D+S+R} & \underline{60.21} & 61.52 & \underline{54.43} & \underline{56.30} & \textbf{48.01} & \textbf{51.39} \\
\midrule
\multirow{8}{*}{\rotatebox[origin=c]{90}{\textbf{\textit{Qwen2.5-72B}}}} & DP & 57.77 & 64.55 & 38.47 & 42.89 & 25.50 & 35.97 \\
& PoT & 39.31 & 41.53 & 50.02 & 51.84 & 29.80 & 32.16 \\
& TCoT & 69.45 & 73.49 & 58.23 & 62.39 & 50.00 & 57.32 \\
& SCoT & 73.15 & 77.21 & 47.75 & 52.07 & 36.42 & 44.93 \\
& \method\textsubscript{R} & 59.19 & 60.50 & 56.95 & 59.28 & 50.66 & 53.97 \\
& \method\textsubscript{D+R} & 56.54 & 57.74 & 57.15 & 59.68 & 51.32 & 54.76 \\
& \method\textsubscript{S+R} & \underline{82.62} & \underline{84.95} & \underline{60.56} & \underline{63.22} & \textbf{63.25} & \textbf{68.21} \\
& \method\textsubscript{D+S+R} & \textbf{83.19} & \textbf{85.39} & \textbf{61.44} & \textbf{64.28} & \underline{60.26} & \underline{64.63} \\
\midrule
% \rowcolor[gray]{.9} 
% \multicolumn{7}{c}{\textbf{\textit{Llama-3.3-70B-Instruct}}} \\ 
% \midrule
% DP & 38.39 & 45.60 & 42.55 & 46.53 & 22.85 & 32.21 \\
% PoT & 31.24 & 33.13 & 52.99 & 54.61 & 23.51 & 25.16 \\
% CoT & 44.76 & 50.68 & 55.16 & 59.46 & 35.76 & 42.73 \\
% \method\textsubscript{R} & 22.34 & 24.11 & 58.23 & 59.99 & 47.02 & 49.86 \\
% \method\textsubscript{D+R} & 22.20 & 23.93 & 57.92 & 59.81 & 44.37 & 47.41 \\
% \method\textsubscript{S+R} & \textbf{57.80} & \textbf{61.60} & \textbf{63.19} & \textbf{65.11} & \textbf{62.58} & \textbf{66.95} \\
% \method\textsubscript{D+S+R} & \underline{57.60} & \underline{61.29} & \underline{62.99} & \underline{65.10} & \underline{62.25} & \underline{65.86} \\
% \midrule
\bottomrule
\end{tabular}
}
\caption{Ablation Study on TAT-QA, TableBench, and CALTAB151; Query Decomposer (D), Table Sanitizer (S), and PoT-based Reasoner (R) are abbreviated as shown; \textbf{Bold} indicates the best performance under the same dataset and model with different Prompts. \underline{Underline} indicates the second-best performance under the same conditions; Qwen2.5-7B~\citep{yang2024qwen2}, Qwen2.5-Code-7B~\citep{hui2024qwen2}, and Qwen2.5-72B~\citep{yang2024qwen2} use their respective instruct-tuning models.}
  \label{tab:ablationTable}
  % \vspace{-10px}
\end{table*}

\subsection{Experimental Settings}

\textbf{Baselines}. We implement the following baselines: 

\begin{itemize}
    \item \textbf{Pre-trained Models} We have selected TAPEX~\citep{liu2022tapex}, and OmniTab~\citep{jiang-etal-2022-omnitab}. The tapex-large-finetuned-wtq and omnitab-large-finetuned-wtq as backbone. Both are fine-tuned on the WikiTQ dataset~\citep{pasupat-liang-2015-compositional}, which is a dataset in the TQA task.
    \item \textbf{Fine-tuning LLMs} We evaluate several fine-tuning LLMs, including TableLlama~\citep{zhang-etal-2024-tablellama}, TableLLM~\citep{zhang2024tablellmenablingtabulardata}, TableGPT2~\citep{su2024tablegpt2largemultimodalmodel}.
    \item \textbf{Prompt-based LLMs} We compare different only prompts methods with \method, include DP, TCoT~\citep{10.5555/3600270.3602070}, PoT~\citep{chen2023program} and SCoT. In addition, we evaluate \method against latest prompt-based methods that include Chain-of-Table~\citep{wang2024chainoftable}, TabSQLify~\citep{nahid-rafiei-2024-tabsqlify}, E5~\citep{zhang-etal-2024-e5}, NormTab~\citep{nahid-rafiei-2024-normtab}, and ReAaTable~\citep{DBLP:journals/corr/abs-2310-00815}\\footnote{The prompts can be found in the Appendix~\ref{appendix:experimentsPrompt}.}.
\end{itemize}

\begin{table*}
\centering
\resizebox{\textwidth}{!}{%
\begin{tabular}{c|c|cc|cc|cc}
\toprule
 \multirow{2}{*}{\textbf{Model}} & \multirow{2}{*}{\textbf{Method}} & \multicolumn{2}{c}{\textbf{TAT-QA}} & \multicolumn{2}{c}{\textbf{TableBench}} & \multicolumn{2}{c}{\textbf{\benchmark}} \\
 \multirow{-1}{*}{} &  & Acc & ROUGE-L & Acc & ROUGE-L & Acc & ROUGE-L \\
\midrule
\multirow{5}{*}{\rotatebox[origin=c]{90}{\textbf{\textit{GPT-4o}}}} & DP & 55.31 & 61.95 & 47.99 & 52.90 & 37.09 & 45.98 \\
& PoT & 48.66 & 49.51 & 57.31 & 59.70 & 45.36 & 49.15 \\
& TCoT & 60.21 & 65.55 & \underline{61.44} & \underline{65.19} & \underline{45.70} & \underline{54.13} \\
& SCoT & \underline{63.97} & \underline{69.27} & 55.40 & 59.90 & 41.39 & 49.32 \\
& \method\textsubscript{D+S+R} & \textbf{80.96} & \textbf{83.24} & \textbf{62.89} & \textbf{65.27} & \textbf{62.25} & \textbf{66.46} \\
\midrule
\multirow{5}{*}{\rotatebox[origin=c]{90}{\textbf{\textit{DeepSeek-V3}}}} & DP & 62.21 & 66.79 & 47.46 & 52.49 & 34.77 & 43.50 \\
& PoT & 57.53 & 58.54 & 49.14 & 51.70 & 37.42 & 40.44 \\
& TCoT & 78.95 & 81.64 & \underline{59.11} & \underline{62.77} & \underline{52.65} & \underline{59.61} \\
& SCoT & \underline{79.66} & \underline{83.62} & 52.43 & 56.85 & 38.74 & 47.69 \\
& \method\textsubscript{D+S+R} & \textbf{83.67} & \textbf{86.13} & \textbf{63.84} & \textbf{66.48} & \textbf{61.92} & \textbf{67.09} \\
\midrule
\bottomrule
\end{tabular}
}
\caption{Transferability of \method: Performance Improvements on GPT-4o~\citep{achiam2023gpt} and DeepSeek-V3~\citep{liu2024deepseek}; \textbf{Bold} indicates the best performance under the same dataset and model with different Prompts. \underline{Underline} indicates the second-best performance under the same conditions.}
  \label{tab:53_transferablity}
  % \vspace{-10px}
\end{table*}

% \subsection{Datasets}

\textbf{Datasets}. We select three TQA datasets for experiments~\footnote{Examples from each dataset can be found in Appendix~\ref{appendix:cases_dataset}.}: 

\begin{itemize}
    \item \textbf{TableBench}~\citep{wu2024tablebench} covers complex TQA questions. The questions involve complex numerical reasoning, fact-checking, Data Analysis, and Visualization. For our experiments, we choose questions related to fact verification and numerical reasoning, containing 493 samples, as these tasks align closely with our research objectives.
    \item \textbf{TAT-QA}~\citep{zhu-etal-2021-tat} challenges models to perform numerical reasoning requiring arithmetic operations, comparisons, and compositional logic. The raw TAT-QA test set contains questions related to tables, table+text (relevant paragraphs), and pure text (relevant paragraphs). Since our task is only related to tables, we select a total of 736 examples where the ``answer\_from'' field is ``table''. 
    \item \textbf{\benchmark} includes multi-hop questions annotated by professional annotators. These questions require reasoning over multiple table entries. It includes 151 table samples.
\end{itemize}

% \footnote{The filename is ``tatqa\_dataset\_test\_gold.json'' in \href{https://github.com/NExTplusplus/TAT-QA/blob/master/dataset_raw/tatqa_dataset_test_gold.json}{Code}.}

% \subsection{LLMs}

% In our experiments, we evaluate our methods with LLMs' sizes are about 7B and 72B, including open-source LLms and close-source LLMs. For open-source LLMs, we evaluate Llama3s~\citep{dubey2024llama}, Qwen2.5s~\citep{yang2024qwen2}, and Qwen2.5-Coder~\citep{hui2024qwen2}. For closed-source LLMs, we evaluate GPT-4o~\citep{achiam2023gpt}.

% \subsection{Experimental Details}

\textbf{Experimental Details.} We configure the LLMs with a maximum token length of 4096 and a temperature of 0.1. For fine-tuning approaches: 

\begin{itemize}
    \item \textbf{TableLlama} uses \textbf{Llama-2-7b-longlora-8k-ft} as its backbone~\citep{touvron2023llama}.
    \item \textbf{TableGPT2} leverages \textbf{Qwen/Qwen2.5-7B}~\citep{yang2024qwen2}.
    \item \textbf{TableLLM-13b} is based on \textbf{CodeLlama-13b-Instruct-hf}~\citep{roziere2023code}.
\end{itemize}

We adopt \textbf{Qwen2.5-7B~\citep{yang2024qwen2}} as the backbone for all prompt-based LLM configurations to ensure a fair comparison.

\textbf{Evaluation Metrics.} We employ accuracy and ROUGE-L~\citep{lin-2004-rouge} as our primary evaluation metrics. Although TableBench uses ROUGE-L as its main metric, we argue that measuring textual overlap alone may not fully capture a model’s performance in complex numerical reasoning. Thus, we include accuracy to more robustly evaluate whether each predicted answer is correct, offering a comprehensive view of the models’ capabilities.

% \input{Tables/SimpleAblationTable}

% \section{Experimental Results}

% Table~\ref{tab:ResultsTables} summarizes the main results across three datasets. In addition, to further analyze the performance of our method, our ablation experiments in Table~\ref{tab:ablationTable} demonstrate the impact of different agents in \method on TQA performance. Finally, to validate the transferability of our approach, experiments in Table~\ref{tab:53_transferablity} reveal high performance improvements when applying the \method to leading large-scale models including GPT-4o and DeepSeek-V3. Our framework is not constrained to specific base models, exhibiting strong transferability across different LLMs.

\subsection{Results and Analysis}

As shown in Table~\ref{tab:ResultsTables}, \method achieves state-of-the-art performance across all three numerical reasoning TQA benchmarks. Notably, the 7B-parameter model even surpasses TableLLM-13B in both accuracy and ROUGE-L, highlighting the effectiveness of \method regardless of model size.

When comparing model categories, we make the following observations: (1) Pre-trained models exhibit the weakest performance. (2) Fine-tuned LLMs rank second overall but struggle significantly on unseen datasets due to biases arising from dataset dependency; for example, TableGPT2-7B and TableLLM-13B@PoT excel on TableBench but show sharp performance drops on the other two datasets. (3) Prompt-based LLMs generally outperform fine-tuned models, suggesting that large models already possess inherent numerical reasoning abilities.

Within the prompt-based category, \method outperforms existing techniques for two key reasons: (1) \textbf{Effective Multi-hop Decomposition}. Competing methods often tackle multi-hop questions in a single pass, which can lead to errors or omissions. (2) \textbf{Robust Table Sanitization.} Many methods rely on SQL-based splitting to handle large tables but overlook unclean cell content, causing calculation errors. By contrast, \method’s dedicated Decomposer and Sanitizer agents ensure higher reliability. Detailed case studies are provided in Appendix~\ref{appendix:CaseStudy}.

Moreover, nearly every method achieves its lowest performance on \benchmark, which we attribute to the absence of data leakage—an advantage that may exist in publicly available datasets. Consequently, \benchmark\ offers a more stringent test of genuine numerical reasoning capabilities.

\subsection{Ablation Study}

Table~\ref{tab:ablationTable} highlights how each agent contributes to final performance. Following \citet{mirzadeh2024gsm}, which suggests that LLMs can accurately compute numerical values via \textbf{R (PoT)}, we include the reasoner (R) in all ablation settings.

\textbf{Effect of the Sanitizer Agent (\method\textsubscript{S})}. Comparing \method\textsubscript{S+R} with \method\textsubscript{R} alone, we observe consistent performance gains—indicating that table sanitization improves data quality for downstream computations. Moreover, the combined \method\textsubscript{D+S+R} setting achieves the highest accuracy overall, underscoring how integrating the Decomposer and Sanitizer agents yields further benefits.

\textbf{Effect of the Decomposer Agent (\method\textsubscript{D})}. In some model–dataset combinations, \method\textsubscript{D+R} slightly underperforms \method\textsubscript{R}. We attribute this to the chaotic nature of certain tables (e.g., complex multi-level headers), which can dilute the value of decomposing the question first. Nonetheless, in most cases, adding the Decomposer (\method\textsubscript{D+R}) or both Decomposer and Sanitizer (\method\textsubscript{D+S+R}) improves performance over \method\textsubscript{R} alone, confirming the overall positive impact of multi-hop question decomposition.

\subsection{Transferability}

Our optimization strategy is fundamentally prompt-based, raising a concern that it only benefits LLMs with limited reasoning capabilities. To test its transferability, we applied \method\ to GPT-4o\citep{achiam2023gpt} and DeepSeek-V3~\citep{liu2024deepseek}, two LLMs widely regarded for their strong reasoning abilities. As shown in Table~\ref{tab:53_transferablity}, our method continues to enhance performance on these powerful models, suggesting that \method\ is not merely compensating for weaker LLMs but provides genuine improvements in numerical reasoning.

Comparing Qwen2.5-72B in Table~\ref{tab:ablationTable} with GPT-4o and DeepSeek-V3 in Table~\ref{tab:53_transferablity}, we find that their results are closely aligned. Although GPT-4o and DeepSeek-V3 have more parameters than Qwen2.5-72B, they also demonstrate comparable outcomes on various public benchmarks~\citep{hendrycks2020measuring,hendrycksmath2021}. These findings indicate that our method effectively boosts complex numerical reasoning performance in LLMs that already possess a robust baseline of numerical reasoning skills.

\section{Conclusion}

We introduced \method, a three-agent, prompt-based framework that significantly elevates numerical reasoning in Table Question Answering (TQA). Our method consistently outperforms pre-trained models, fine-tuned LLMs, and other prompt-based solutions, demonstrating its effectiveness in handling complex tabular data. By decomposing multi-hop questions and sanitizing noisy table content, \method\ fully harnesses the inherent numerical reasoning capabilities of LLMs, enhancing their performance regardless of parameter size.

This work also has practical implications for various real-world domains—such as finance, business intelligence, healthcare, and e-commerce—where robust analysis of complex, noisy tabular data is critical. The prompt-based nature of \method\ lowers barriers to adoption by reducing the need for extensive data annotation or specialized training, enabling more cost-effective and scalable deployment of powerful TQA systems. Moreover, our framework’s transferability across diverse model architectures highlights its potential for broader integration into enterprise workflows, data analytics platforms, and decision-support systems that require reliable, multi-step numerical calculations over large datasets. We hope these findings inspire further research and innovation in complex numerical reasoning for TQA, spurring the development of even more versatile and efficient solutions.

\section*{Limitations}

Although the \method performs well, it still falls far short of human performance in answering complex tabular numerical reasoning questions. Our method is prompt-only, and its performance is limited by the reasoning ability of the LLM.

\benchmark requires manual verification during the final validation stage, which makes the annotation process costly. As a result, the dataset size is relatively small, and the range of question types is limited. In future work, we plan to expand the dataset by increasing its size, enriching the diversity of question types, and covering a broader range of domains to enable more comprehensive evaluations.

We opted for a restricted Decomposer that operates solely on the question. While this design choice may slightly degrade performance in some isolated cases, our experiments show that it brings consistent and often mild improvements in the majority of scenarios, especially in terms of stability and generalizability across datasets. We acknowledge the trade-off here and consider it a pragmatic decision to ensure robustness and utility in real-world settings. In future work, we plan to explore hybrid approaches that can selectively incorporate table signals while preserving decomposition reliability.

The Sanitizer is constrained by the reflection capabilities of the underlying base model. In some cases, even after multiple iterations, it fails to correctly repair the tables, leading to excessive and redundant calls. Moving forward, we will implement call monitoring and fallback mechanisms to ensure that when the Sanitizer fails, a default resolution strategy can be applied efficiently.

\section*{Ethical Considerations}

This work involves the use of AI systems in two aspects. First, AI tools were utilized to assist with the translation of this paper. Second, AI models were employed during the data construction process; however, all generated data strictly followed the guidelines described in Appendix~\ref{appendix:License} and was used solely for academic and research purposes.

We ensured that no personally identifiable information (PII) or sensitive data was included in \benchmark.

\section*{Acknowledgments}

This research was supported by the National Key Research and Development Program of China (Subject No.2022YFC3302904) and the Young Scientists Fund of the National Natural Science Foundation of China (Grant No.72304215).

% This document has been adapted
% by Steven Bethard, Ryan Cotterell and Rui Yan
% from the instructions for earlier ACL and NAACL proceedings, including those for
% ACL 2019 by Douwe Kiela and Ivan Vuli\'{c},
% NAACL 2019 by Stephanie Lukin and Alla Roskovskaya,
% ACL 2018 by Shay Cohen, Kevin Gimpel, and Wei Lu,
% NAACL 2018 by Margaret Mitchell and Stephanie Lukin,
% Bib\TeX{} suggestions for (NA)ACL 2017/2018 from Jason Eisner,
% ACL 2017 by Dan Gildea and Min-Yen Kan,
% NAACL 2017 by Margaret Mitchell,
% ACL 2012 by Maggie Li and Michael White,
% ACL 2010 by Jing-Shin Chang and Philipp Koehn,
% ACL 2008 by Johanna D. Moore, Simone Teufel, James Allan, and Sadaoki Furui,
% ACL 2005 by Hwee Tou Ng and Kemal Oflazer,
% ACL 2002 by Eugene Charniak and Dekang Lin,
% and earlier ACL and EACL formats written by several people, including
% John Chen, Henry S. Thompson and Donald Walker.
% Additional elements were taken from the formatting instructions of the \emph{International Joint Conference on Artificial Intelligence} and the \emph{Conference on Computer Vision and Pattern Recognition}.

% \bibliographystyle{acl_natbib}
\bibliography{custom}

@article{wu2024tablebench,
  title={TableBench: A Comprehensive and Complex Benchmark for Table Question Answering},
  author={Wu, Xianjie and Yang, Jian and Chai, Linzheng and Zhang, Ge and Liu, Jiaheng and Du, Xinrun and Liang, Di and Shu, Daixin and Cheng, Xianfu and Sun, Tianzhen and others},
  journal={arXiv preprint arXiv:2408.09174},
  year={2024}
}

@inproceedings{chen2021finqa,
  title={FinQA: A Dataset of Numerical Reasoning over Financial Data},
  author={Chen, Zhiyu and Chen, Wenhu and Smiley, Charese and Shah, Sameena and Borova, Iana and Langdon, Dylan and Moussa, Reema and Beane, Matt and Huang, Ting-Hao and Routledge, Bryan R and others},
  booktitle={Proceedings of the 2021 Conference on Empirical Methods in Natural Language Processing},
  pages={3697--3711},
  year={2021}
}

@article{mirzadeh2024gsm,
  title={Gsm-symbolic: Understanding the limitations of mathematical reasoning in large language models},
  author={Mirzadeh, Iman and Alizadeh, Keivan and Shahrokhi, Hooman and Tuzel, Oncel and Bengio, Samy and Farajtabar, Mehrdad},
  journal={arXiv preprint arXiv:2410.05229},
  year={2024}
}

@inproceedings{lu2023dynamic,
    title={Dynamic Prompt Learning via Policy Gradient for Semi-structured Mathematical Reasoning},
    author={Lu, Pan and Qiu, Liang and Chang, Kai-Wei and Wu, Ying Nian and Zhu, Song-Chun and Rajpurohit, Tanmay and Clark, Peter and Kalyan, Ashwin},
    booktitle={International Conference on Learning Representations (ICLR)},
    year={2023}
}

@inproceedings{lin-2004-rouge,
    title = "{ROUGE}: A Package for Automatic Evaluation of Summaries",
    author = "Lin, Chin-Yew",
    booktitle = "Text Summarization Branches Out",
    month = jul,
    year = "2004",
    address = "Barcelona, Spain",
    publisher = "Association for Computational Linguistics",
    url = "https://aclanthology.org/W04-1013/",
    pages = "74--81"
}

@inproceedings{zhang-etal-2024-e5,
    title = "$E^5$: Zero-shot Hierarchical Table Analysis using Augmented {LLM}s via Explain, Extract, Execute, Exhibit and Extrapolate",
    author = "Zhang, Zhehao  and
      Gao, Yan  and
      Lou, Jian-Guang",
    editor = "Duh, Kevin  and
      Gomez, Helena  and
      Bethard, Steven",
    booktitle = "Proceedings of the 2024 Conference of the North American Chapter of the Association for Computational Linguistics: Human Language Technologies (Volume 1: Long Papers)",
    month = jun,
    year = "2024",
    address = "Mexico City, Mexico",
    publisher = "Association for Computational Linguistics",
    url = "https://aclanthology.org/2024.naacl-long.68/",
    doi = "10.18653/v1/2024.naacl-long.68",
    pages = "1244--1258"
}

@inproceedings{zhu-etal-2021-tat,
    title = "{TAT}-{QA}: A Question Answering Benchmark on a Hybrid of Tabular and Textual Content in Finance",
    author = "Zhu, Fengbin  and
      Lei, Wenqiang  and
      Huang, Youcheng  and
      Wang, Chao  and
      Zhang, Shuo  and
      Lv, Jiancheng  and
      Feng, Fuli  and
      Chua, Tat-Seng",
    editor = "Zong, Chengqing  and
      Xia, Fei  and
      Li, Wenjie  and
      Navigli, Roberto",
    booktitle = "Proceedings of the 59th Annual Meeting of the Association for Computational Linguistics and the 11th International Joint Conference on Natural Language Processing (Volume 1: Long Papers)",
    month = aug,
    year = "2021",
    address = "Online",
    publisher = "Association for Computational Linguistics",
    url = "https://aclanthology.org/2021.acl-long.254/",
    doi = "10.18653/v1/2021.acl-long.254",
    pages = "3277--3287"
}

@inproceedings{katsis-etal-2022-ait,
    title = "{AIT-QA}: {Q}uestion Answering Dataset over Complex Tables in the Airline Industry",
    author = "Katsis, Yannis  and
      Chemmengath, Saneem  and
      Kumar, Vishwajeet  and
      Bharadwaj, Samarth  and
      Canim, Mustafa  and
      Glass, Michael  and
      Gliozzo, Alfio  and
      Pan, Feifei  and
      Sen, Jaydeep  and
      Sankaranarayanan, Karthik  and
      Chakrabarti, Soumen",
    editor = "Loukina, Anastassia  and
      Gangadharaiah, Rashmi  and
      Min, Bonan",
    booktitle = "Proceedings of the 2022 Conference of the North American Chapter of the Association for Computational Linguistics: Human Language Technologies: Industry Track",
    month = jul,
    year = "2022",
    address = "Hybrid: Seattle, Washington + Online",
    publisher = "Association for Computational Linguistics",
    url = "https://aclanthology.org/2022.naacl-industry.34/",
    doi = "10.18653/v1/2022.naacl-industry.34",
    pages = "305--314"
}

@article{
    chen2023program,
    title={Program of Thoughts Prompting: Disentangling Computation from Reasoning for Numerical Reasoning Tasks},
    author={Wenhu Chen and Xueguang Ma and Xinyi Wang and William W. Cohen},
    journal={Transactions on Machine Learning Research},
    issn={2835-8856},
    year={2023},
    url={https://openreview.net/forum?id=YfZ4ZPt8zd},
    note={}
}

@inproceedings{10.1145/3539618.3591708,
author = {Ye, Yunhu and Hui, Binyuan and Yang, Min and Li, Binhua and Huang, Fei and Li, Yongbin},
title = {Large Language Models are Versatile Decomposers: Decomposing Evidence and Questions for Table-based Reasoning},
year = {2023},
isbn = {9781450394086},
publisher = {Association for Computing Machinery},
address = {New York, NY, USA},
url = {https://doi.org/10.1145/3539618.3591708},
doi = {10.1145/3539618.3591708},
booktitle = {Proceedings of the 46th International ACM SIGIR Conference on Research and Development in Information Retrieval},
pages = {174–184},
numpages = {11},
location = {Taipei, Taiwan},
series = {SIGIR '23}
}

@article{badaro-etal-2023-transformers,
    title = "Transformers for Tabular Data Representation: A Survey of Models and Applications",
    author = "Badaro, Gilbert  and
      Saeed, Mohammed  and
      Papotti, Paolo",
    journal = "Transactions of the Association for Computational Linguistics",
    volume = "11",
    year = "2023",
    address = "Cambridge, MA",
    publisher = "MIT Press",
    url = "https://aclanthology.org/2023.tacl-1.14/",
    doi = "10.1162/tacl_a_00544",
    pages = "227--249"
}

@inproceedings{zhang-etal-2024-tablellama,
    title = "{T}able{L}lama: Towards Open Large Generalist Models for Tables",
    author = "Zhang, Tianshu  and
      Yue, Xiang  and
      Li, Yifei  and
      Sun, Huan",
    editor = "Duh, Kevin  and
      Gomez, Helena  and
      Bethard, Steven",
    booktitle = "Proceedings of the 2024 Conference of the North American Chapter of the Association for Computational Linguistics: Human Language Technologies (Volume 1: Long Papers)",
    month = jun,
    year = "2024",
    address = "Mexico City, Mexico",
    publisher = "Association for Computational Linguistics",
    url = "https://aclanthology.org/2024.naacl-long.335/",
    doi = "10.18653/v1/2024.naacl-long.335",
    pages = "6024--6044"
}

@article{10.1145/3654979,
author = {Li, Peng and He, Yeye and Yashar, Dror and Cui, Weiwei and Ge, Song and Zhang, Haidong and Rifinski Fainman, Danielle and Zhang, Dongmei and Chaudhuri, Surajit},
title = {Table-GPT: Table Fine-tuned GPT for Diverse Table Tasks},
year = {2024},
issue_date = {June 2024},
publisher = {Association for Computing Machinery},
address = {New York, NY, USA},
volume = {2},
number = {3},
url = {https://doi.org/10.1145/3654979},
doi = {10.1145/3654979},
journal = {Proc. ACM Manag. Data},
month = may,
articleno = {176},
numpages = {28}
}

@inproceedings{10.5555/3600270.3602070,
author = {Wei, Jason and Wang, Xuezhi and Schuurmans, Dale and Bosma, Maarten and Ichter, Brian and Xia, Fei and Chi, Ed H. and Le, Quoc V. and Zhou, Denny},
title = {Chain-of-thought prompting elicits reasoning in large language models},
year = {2024},
isbn = {9781713871088},
publisher = {Curran Associates Inc.},
address = {Red Hook, NY, USA},
booktitle = {Proceedings of the 36th International Conference on Neural Information Processing Systems},
articleno = {1800},
numpages = {14},
location = {New Orleans, LA, USA},
series = {NIPS '22}
}

@article{kojima2022large,
  title={Large language models are zero-shot reasoners},
  author={Kojima, Takeshi and Gu, Shixiang Shane and Reid, Machel and Matsuo, Yutaka and Iwasawa, Yusuke},
  journal={Advances in neural information processing systems},
  volume={35},
  pages={22199--22213},
  year={2022}
}

@inproceedings{nahid-rafiei-2024-tabsqlify,
    title = "{T}ab{SQL}ify: Enhancing Reasoning Capabilities of {LLM}s Through Table Decomposition",
    author = "Nahid, Md  and
      Rafiei, Davood",
    editor = "Duh, Kevin  and
      Gomez, Helena  and
      Bethard, Steven",
    booktitle = "Proceedings of the 2024 Conference of the North American Chapter of the Association for Computational Linguistics: Human Language Technologies (Volume 1: Long Papers)",
    month = jun,
    year = "2024",
    address = "Mexico City, Mexico",
    publisher = "Association for Computational Linguistics",
    url = "https://aclanthology.org/2024.naacl-long.320/",
    doi = "10.18653/v1/2024.naacl-long.320",
    pages = "5725--5737"
}

@inproceedings{deng-etal-2024-investigating,
    title = "Investigating Data Contamination in Modern Benchmarks for Large Language Models",
    author = "Deng, Chunyuan  and
      Zhao, Yilun  and
      Tang, Xiangru  and
      Gerstein, Mark  and
      Cohan, Arman",
    editor = "Duh, Kevin  and
      Gomez, Helena  and
      Bethard, Steven",
    booktitle = "Proceedings of the 2024 Conference of the North American Chapter of the Association for Computational Linguistics: Human Language Technologies (Volume 1: Long Papers)",
    month = jun,
    year = "2024",
    address = "Mexico City, Mexico",
    publisher = "Association for Computational Linguistics",
    url = "https://aclanthology.org/2024.naacl-long.482/",
    doi = "10.18653/v1/2024.naacl-long.482",
    pages = "8706--8719"
}

@inproceedings{
liu2022tapex,
title={{TAPEX}: Table Pre-training via Learning a Neural {SQL} Executor},
author={Qian Liu and Bei Chen and Jiaqi Guo and Morteza Ziyadi and Zeqi Lin and Weizhu Chen and Jian-Guang Lou},
booktitle={International Conference on Learning Representations},
year={2022},
url={https://openreview.net/forum?id=O50443AsCP}
}

@inproceedings{herzig-etal-2020-tapas,
    title = "{T}a{P}as: Weakly Supervised Table Parsing via Pre-training",
    author = {Herzig, Jonathan  and
      Nowak, Pawel Krzysztof  and
      M{\"u}ller, Thomas  and
      Piccinno, Francesco  and
      Eisenschlos, Julian},
    editor = "Jurafsky, Dan  and
      Chai, Joyce  and
      Schluter, Natalie  and
      Tetreault, Joel",
    booktitle = "Proceedings of the 58th Annual Meeting of the Association for Computational Linguistics",
    month = jul,
    year = "2020",
    address = "Online",
    publisher = "Association for Computational Linguistics",
    url = "https://aclanthology.org/2020.acl-main.398/",
    doi = "10.18653/v1/2020.acl-main.398",
    pages = "4320--4333"
}

@inproceedings{jiang-etal-2022-omnitab,
  title = "{O}mni{T}ab: Pretraining with Natural and Synthetic Data for Few-shot Table-based Question Answering",
  author = "Jiang, Zhengbao and Mao, Yi and He, Pengcheng and Neubig, Graham and Chen, Weizhu",
  booktitle = "Proceedings of the 2022 Conference of the North American Chapter of the Association for Computational Linguistics: Human Language Technologies",
  month = jul,
  year = "2022",
}

@inproceedings{
wang2024chainoftable,
title={Chain-of-Table: Evolving Tables in the Reasoning Chain for Table Understanding},
author={Zilong Wang and Hao Zhang and Chun-Liang Li and Julian Martin Eisenschlos and Vincent Perot and Zifeng Wang and Lesly Miculicich and Yasuhisa Fujii and Jingbo Shang and Chen-Yu Lee and Tomas Pfister},
booktitle={The Twelfth International Conference on Learning Representations},
year={2024},
url={https://openreview.net/forum?id=4L0xnS4GQM}
}

@inproceedings{nahid-rafiei-2024-normtab,
    title = "{N}orm{T}ab: Improving Symbolic Reasoning in {LLM}s Through Tabular Data Normalization",
    author = "Nahid, Md Mahadi Hasan and
      Rafiei, Davood",
    editor = "Al-Onaizan, Yaser  and
      Bansal, Mohit  and
      Chen, Yun-Nung",
    booktitle = "Findings of the Association for Computational Linguistics: EMNLP 2024",
    month = nov,
    year = "2024",
    address = "Miami, Florida, USA",
    publisher = "Association for Computational Linguistics",
    url = "https://aclanthology.org/2024.findings-emnlp.203",
    pages = "3569--3585",
}

@article{yang2024qwen2,
  title={Qwen2.5 Technical Report},
  author={Yang, An and Yang, Baosong and Zhang, Beichen and Hui, Binyuan and Zheng, Bo and Yu, Bowen and Li, Chengyuan and Liu, Dayiheng and Huang, Fei and Wei, Haoran and others},
  journal={arXiv preprint arXiv:2412.15115},
  year={2024}
}

@article{hui2024qwen2,
  title={Qwen2.5-coder technical report},
  author={Hui, Binyuan and Yang, Jian and Cui, Zeyu and Yang, Jiaxi and Liu, Dayiheng and Zhang, Lei and Liu, Tianyu and Zhang, Jiajun and Yu, Bowen and Lu, Keming and others},
  journal={arXiv preprint arXiv:2409.12186},
  year={2024}
}

@article{liu2024deepseek,
  title={Deepseek-v3 technical report},
  author={Liu, Aixin and Feng, Bei and Xue, Bing and Wang, Bingxuan and Wu, Bochao and Lu, Chengda and Zhao, Chenggang and Deng, Chengqi and Zhang, Chenyu and Ruan, Chong and others},
  journal={arXiv preprint arXiv:2412.19437},
  year={2024}
}

@article{achiam2023gpt,
  title={Gpt-4 technical report},
  author={Achiam, Josh and Adler, Steven and Agarwal, Sandhini and Ahmad, Lama and Akkaya, Ilge and Aleman, Florencia Leoni and Almeida, Diogo and Altenschmidt, Janko and Altman, Sam and Anadkat, Shyamal and others},
  journal={arXiv preprint arXiv:2303.08774},
  year={2023}
}

@misc{zhang2024tablellmenablingtabulardata,
      title={TableLLM: Enabling Tabular Data Manipulation by LLMs in Real Office Usage Scenarios}, 
      author={Xiaokang Zhang and Jing Zhang and Zeyao Ma and Yang Li and Bohan Zhang and Guanlin Li and Zijun Yao and Kangli Xu and Jinchang Zhou and Daniel Zhang-Li and Jifan Yu and Shu Zhao and Juanzi Li and Jie Tang},
      year={2024},
      eprint={2403.19318},
      archivePrefix={arXiv},
      primaryClass={cs.CL},
      url={https://arxiv.org/abs/2403.19318}, 
}

@misc{su2024tablegpt2largemultimodalmodel,
      title={TableGPT2: A Large Multimodal Model with Tabular Data Integration}, 
      author={Aofeng Su and Aowen Wang and Chao Ye and Chen Zhou and Ga Zhang and Guangcheng Zhu and Haobo Wang and Haokai Xu and Hao Chen and Haoze Li and Haoxuan Lan and Jiaming Tian and Jing Yuan and Junbo Zhao and Junlin Zhou and Kaizhe Shou and Liangyu Zha and Lin Long and Liyao Li and Pengzuo Wu and Qi Zhang and Qingyi Huang and Saisai Yang and Tao Zhang and Wentao Ye and Wufang Zhu and Xiaomeng Hu and Xijun Gu and Xinjie Sun and Xiang Li and Yuhang Yang and Zhiqing Xiao},
      year={2024},
      eprint={2411.02059},
      archivePrefix={arXiv},
      primaryClass={cs.LG},
      url={https://arxiv.org/abs/2411.02059}, 
}

@inproceedings{liu-etal-2024-rethinking,
    title = "Rethinking Tabular Data Understanding with Large Language Models",
    author = "Liu, Tianyang  and
      Wang, Fei  and
      Chen, Muhao",
    editor = "Duh, Kevin  and
      Gomez, Helena  and
      Bethard, Steven",
    booktitle = "Proceedings of the 2024 Conference of the North American Chapter of the Association for Computational Linguistics: Human Language Technologies (Volume 1: Long Papers)",
    month = jun,
    year = "2024",
    address = "Mexico City, Mexico",
    publisher = "Association for Computational Linguistics",
    url = "https://aclanthology.org/2024.naacl-long.26/",
    doi = "10.18653/v1/2024.naacl-long.26",
    pages = "450--482"
}

@inproceedings{
yao2023react,
title={ReAct: Synergizing Reasoning and Acting in Language Models},
author={Shunyu Yao and Jeffrey Zhao and Dian Yu and Nan Du and Izhak Shafran and Karthik R Narasimhan and Yuan Cao},
booktitle={The Eleventh International Conference on Learning Representations },
year={2023},
url={https://openreview.net/forum?id=WE_vluYUL-X}
}

@inproceedings{pasupat-liang-2015-compositional,
    title = "Compositional Semantic Parsing on Semi-Structured Tables",
    author = "Pasupat, Panupong  and
      Liang, Percy",
    editor = "Zong, Chengqing  and
      Strube, Michael",
    booktitle = "Proceedings of the 53rd Annual Meeting of the Association for Computational Linguistics and the 7th International Joint Conference on Natural Language Processing (Volume 1: Long Papers)",
    month = jul,
    year = "2015",
    address = "Beijing, China",
    publisher = "Association for Computational Linguistics",
    url = "https://aclanthology.org/P15-1142/",
    doi = "10.3115/v1/P15-1142",
    pages = "1470--1480"
}

@inproceedings{biran-etal-2024-hopping,
    title = "Hopping Too Late: Exploring the Limitations of Large Language Models on Multi-Hop Queries",
    author = "Biran, Eden  and
      Gottesman, Daniela  and
      Yang, Sohee  and
      Geva, Mor  and
      Globerson, Amir",
    editor = "Al-Onaizan, Yaser  and
      Bansal, Mohit  and
      Chen, Yun-Nung",
    booktitle = "Proceedings of the 2024 Conference on Empirical Methods in Natural Language Processing",
    month = nov,
    year = "2024",
    address = "Miami, Florida, USA",
    publisher = "Association for Computational Linguistics",
    url = "https://aclanthology.org/2024.emnlp-main.781/",
    doi = "10.18653/v1/2024.emnlp-main.781",
    pages = "14113--14130"
}

@inproceedings{guan2024mfort,
  title={Mfort-qa: Multi-hop few-shot open rich table question answering},
  author={Guan, Che and Huang, Mengyu and Zhang, Peng},
  booktitle={Proceedings of the 2024 10th International Conference on Computing and Artificial Intelligence},
  pages={434--442},
  year={2024}
}

@article{touvron2023llama,
  title={Llama 2: Open foundation and fine-tuned chat models},
  author={Touvron, Hugo and Martin, Louis and Stone, Kevin and Albert, Peter and Almahairi, Amjad and Babaei, Yasmine and Bashlykov, Nikolay and Batra, Soumya and Bhargava, Prajjwal and Bhosale, Shruti and others},
  journal={arXiv preprint arXiv:2307.09288},
  year={2023}
}

@article{roziere2023code,
  title={Code llama: Open foundation models for code},
  author={Roziere, Baptiste and Gehring, Jonas and Gloeckle, Fabian and Sootla, Sten and Gat, Itai and Tan, Xiaoqing Ellen and Adi, Yossi and Liu, Jingyu and Sauvestre, Romain and Remez, Tal and others},
  journal={arXiv preprint arXiv:2308.12950},
  year={2023}
}

@inproceedings{perez2020unsupervised,
    title={Unsupervised Question Decomposition for Question Answering},
    author={Ethan Perez and Patrick Lewis and Wen-tau Yih and Kyunghyun Cho and Douwe Kiela},
    year={2020},
    booktitle={EMNLP},
    url={https://arxiv.org/abs/2002.09758}
}

@inproceedings{zhang-etal-2019-complex,
    title = "Complex Question Decomposition for Semantic Parsing",
    author = "Zhang, Haoyu  and
      Cai, Jingjing  and
      Xu, Jianjun  and
      Wang, Ji",
    editor = "Korhonen, Anna  and
      Traum, David  and
      M{\`a}rquez, Llu{\'i}s",
    booktitle = "Proceedings of the 57th Annual Meeting of the Association for Computational Linguistics",
    month = jul,
    year = "2019",
    address = "Florence, Italy",
    publisher = "Association for Computational Linguistics",
    url = "https://aclanthology.org/P19-1440/",
    doi = "10.18653/v1/P19-1440",
    pages = "4477--4486"
}

@article{hendrycks2020measuring,
  title={Measuring massive multitask language understanding},
  author={Hendrycks, Dan and Burns, Collin and Basart, Steven and Zou, Andy and Mazeika, Mantas and Song, Dawn and Steinhardt, Jacob},
  journal={arXiv preprint arXiv:2009.03300},
  year={2020}
}

@article{hendrycksmath2021,
  title={Measuring Mathematical Problem Solving With the MATH Dataset},
  author={Dan Hendrycks and Collin Burns and Saurav Kadavath and Akul Arora and Steven Basart and Eric Tang and Dawn Song and Jacob Steinhardt},
  journal={NeurIPS},
  year={2021}
}

@article{DBLP:journals/corr/abs-2310-00815,
  author       = {Yunjia Zhang and
                  Jordan Henkel and
                  Avrilia Floratou and
                  Joyce Cahoon and
                  Shaleen Deep and
                  Jignesh M. Patel},
  title        = {ReAcTable: Enhancing ReAct for Table Question Answering},
  journal      = {CoRR},
  volume       = {abs/2310.00815},
  year         = {2023},
  url          = {https://doi.org/10.48550/arXiv.2310.00815},
  doi          = {10.48550/ARXIV.2310.00815},
  eprinttype    = {arXiv},
  eprint       = {2310.00815},
  bibsource    = {dblp computer science bibliography, https://dblp.org}
}

@article{yuan2025enhancing,
  title={Enhancing Visual Grounding for GUI Agents via Self-Evolutionary Reinforcement Learning},
  author={Yuan, Xinbin and Zhang, Jian and Li, Kaixin and Cai, Zhuoxuan and Yao, Lujian and Chen, Jie and Wang, Enguang and Hou, Qibin and Chen, Jinwei and Jiang, Peng-Tao and others},
  journal={arXiv preprint arXiv:2505.12370},
  year={2025}
}

@misc{zhang2025agentorchestrahierarchicalmultiagentframework,
      title={AgentOrchestra: Orchestrating Hierarchical Multi-Agent Intelligence with the Tool-Environment-Agent(TEA) Protocol}, 
      author={Wentao Zhang and Liang Zeng and Yuzhen Xiao and Yongcong Li and Ce Cui and Yilei Zhao and Rui Hu and Yang Liu and Yahui Zhou and Bo An},
      year={2025},
      eprint={2506.12508},
      archivePrefix={arXiv},
      primaryClass={cs.AI},
      url={https://arxiv.org/abs/2506.12508}, 
}

\appendix

\section{License}
\label{appendix:License}

For all datasets in our experiments, TableBench is under the license of MIT. The AIT-QA dataset is under the license of CDLA-Sharing-1.0. The TAT-QA dataset is under the license of Creative Commons (CC BY) Attribution 4.0 International. The FinQA dataset is under the license of Creative Commons Attribution 4.0 International. All of these licenses allow their data for academic use.

\section{Construction of \benchmark}

\label{appendix:caltab151}

\begin{figure*}[t]
  \includegraphics[width=\textwidth]{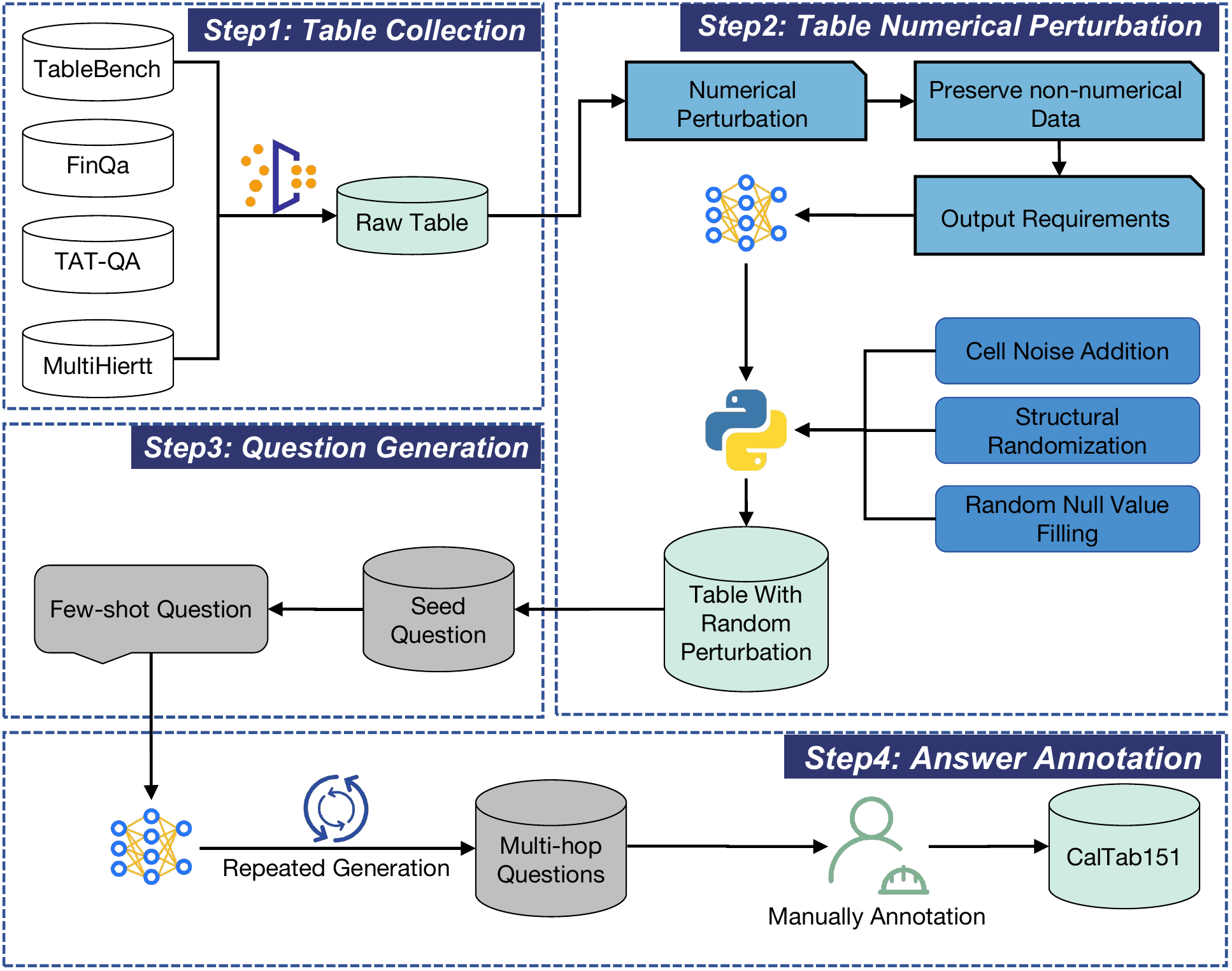}
  \caption{The construction pipeline of \benchmark.}
  \label{fig:CalTab151FrameWork}
\end{figure*}

The pipeline of constructing the \benchmark dataset can refer to Figure~\ref{fig:CalTab151FrameWork}. Figures~\ref{fig:QueryGeneration} and~\ref{fig:NumericalPerturabation} are Prompts used in the construction of \benchmark.

The datasets utilized in this study were sourced exclusively from pre-existing open-source repositories, with strict adherence to their respective licensing agreements throughout both data acquisition and implementation phases, and are solely for academic use.

Manually annotation is carried out by students with a specialized background in computer science. This entire process follows a strict data annotation workflow. We provide professional computing devices to support the annotators in completing the task.

\begin{figure*}[t]
\begin{tcolorbox}[
  colback=white, % 背景颜色为黑色
  colframe=black, % 边框颜色为黑色
  colupper=gray!70, % 标题颜色为浅灰色
  fontupper=\footnotesize, % 内容字体大小
  % fonttitle=\bfseries\footnotesize, % 标题字体
  coltitle=white, % 标题字体颜色
  coltext=black, % 内容字体颜色
  boxrule=0.5mm, % 边框厚度
  title=Prompt Template of Query Generation
]
You are a data augmentation assistant. Your goal is to choose two most relevant SubQueries and integrate two of those SubQueries into a human readable and professional question. The Question should demonstrate a logical progression and incorporate nested relationships.\\
\textbf{\#\# Reference SubQueries}\\

What was the amount of unrecognized stock-based compensation expense related to unvested employee stock options in 2019?\\
What was the total stock-based compensation expense amount in 2018?\\
How long is it expected to take for the unrecognized stock-based compensation expense related to unvested RSUs to be recognized?\\
What is the total stock-based compensation expense and unrecognized stock-based compensation expense in 2019?\\
What was the change in the amount of stock options in 2019 from 2018?\\
What was the percentage change in the amount of RSUs in 2019 from 2018?\\

\textbf{\#\# answer}\\

The questions “What was the change in the amount of stock options in 2019 from 2018?” and “What was the percentage change in the amount of RSUs in 2019 from 2018?” both reference the period “in 2019 from 2018.” Therefore, we can combine them into a single question about the same time period. The final question is formatted in JSON as follows:\\
\begin{verbatim}
```json
{
  "Question": "What was the change in the amount of stock options in 2019 from 2018? Additionally,
  what was the percentage change in the amount of RSUs during the same period?"
}
```
\end{verbatim}

\textbf{\#\# Reference SubQueries}\\
\begin{verbatim}
{{ReferQuestion}}
\end{verbatim}
\textbf{\#\# answer}\\

You must output in json format:\\

- Question: string, a human readable and professional question, consist of two most relevant SubQueries.\\
\begin{verbatim}
```json
{
  "Question": "string, a human readable and professional question."
}
```
\end{verbatim}
Give a final question in json format in the end, Let's think step and step!
\end{tcolorbox}
\caption{\textbf{Query Generation} generates multi-hop queries by few-shot prompt.}
\label{fig:QueryGeneration}
\end{figure*}
\begin{figure*}[t]
\begin{tcolorbox}[
  colback=white, % 背景颜色为黑色
  colframe=black, % 边框颜色为黑色
  colupper=gray!70, % 标题颜色为浅灰色
  fontupper=\footnotesize, % 内容字体大小
  % fonttitle=\bfseries\footnotesize, % 标题字体
  coltitle=white, % 标题字体颜色
  coltext=black, % 内容字体颜色
  boxrule=0.5mm, % 边框厚度
  title=Prompt Template of Numerical Perturbation
]
You are a data augmentation assistant. Your task is to generate a new table by applying the following transformation rules exclusively to numeric data in the table, including numbers stored as strings. Specifically:\\
1. \texttt{Numerical Perturbation}:\\
\hspace*{1em}- Identify Numeric Data:\\
\hspace*{2em}- Include all numeric values (integers, floats, or numbers stored as strings), even if mixed with other characters (e.g., \$100.00, 123.45kg).\\
\hspace*{2em}- Exclude data that clearly represents dates or times (e.g., YYYY-MM-DD, MM/DD/YYYY, or time formats like HH:MM:SS).\\
\hspace*{1em}- Apply Perturbation:\\
\hspace*{2em}- Randomly adjust numeric values (including those within strings) by up to ±3\%-5\% of their original value.\\
\hspace*{1em}- Maintain data realism:\\
\hspace*{2em}- If the original value is an integer, the perturbed value must remain an integer.\\
\hspace*{2em}- If the value is a float, retain its decimal format with appropriate precision.\\
\hspace*{2em}- For strings containing numeric values, only adjust the numeric portion, leaving non-numeric characters intact (e.g., \$100.00 → \$103.00).\\
\hspace*{1em}- Ensure that the perturbation keeps the values realistic within the context of the data.\\
2. \texttt{Preserve Other Data}:\\
\hspace*{1em}- Retain all non-numeric columns, values, and formats unchanged.\\
\hspace*{1em}- Date or time columns must not be perturbed or modified.\\
3. \texttt{Output Requirements}:\\
\hspace*{1em}- Directly output the augmented table in JSON format, maintaining the structure of the input table. The JSON must include:\\
\hspace*{2em}- "columns": An array of column names.\\
\hspace*{2em}- "data": A 2D array of table rows after transformation.\\
\hspace*{2em}- "index": The original index of each row.\\

- \texttt{Identification Criteria}:\\
\hspace*{1em}- Numeric Columns: Include numbers (int, float) and numbers stored as strings, even if they contain additional non-numeric characters.\\
\hspace*{1em}- Date Columns: Avoid perturbation for values matching common date/time formats (e.g., YYYY-MM-DD or HH:MM:SS).\\

\textbf{\#\# Input}
\begin{verbatim}
```json
{{Inputs}}
```
\end{verbatim}
\textbf{\#\# Output}
\end{tcolorbox}
\caption{\textbf{Numerical Perturbation} with zero-shot prompt, generate a new table by applying transformation rules exclusively to numeric data in the table, including numbers stored as strings, but excluding data of dates or times.}
\label{fig:NumericalPerturabation}
\end{figure*}

\section{Examples of Datasets}
\label{appendix:cases_dataset}

% \begin{figure*}
%   \centering
%   \includegraphics[width=\textwidth]{attachments/cases_datasets.pdf}
%   \caption{Cases of TAT-QA, TableBench, and \benchmark.}
%   \label{fig:cases_dataset}
% \end{figure*}

% \usepackage{booktabs}
% \usepackage{graphicx}
{
\begin{table*}[t]
\centering
\resizebox{\textwidth}{!}{%
\begin{tabular}{@{}|l|l|l|l|@{}}
\toprule
\textbf{Table} &
  \multicolumn{1}{c|}{\textbf{Question}} &
  \multicolumn{1}{c|}{\textbf{Answer}} \\ \midrule
  \begin{tabular}[c]{@{}l@{}} \{'columns': \texttt{[} \\ 'Remuneration key performance indicator', '2019 actual', '2019 threshold', \\ '2019 target', '2019 maximum', 'Remuneration measure'\texttt{]}, \\ 'data': \texttt{[} \\ \texttt{[}'Group operating profit (\u00a3m)', '277.3', '256.7', '270.3', '283.8', \\ 'Annual Incentive Plan'\texttt{]}, \\ \texttt{[}'Group cash generation (\u00a3m)', '296.4', '270.7', '285.0', '299.2', \\ 'Annual Incentive Plan'\texttt{]}, \\ \texttt{[}'Group ROCE (\%)', '54.5', '50.1', '52.7', '55.3', 'Annual Incentive Plan'\texttt{]}, \\ \texttt{[}'2017-2019 EPS (\%)', '57.5', '27.6', 'N/A', '52.3', 'Performance Share Plan'\texttt{]}, \\ \texttt{[}'2017-2019 relative TSR (percentile TSR)', \\ '94th', '50th', 'N/A', '75th', 'Performance Share Plan'\texttt{]}\texttt{]}\} \end{tabular} &
  \begin{tabular}[c]{@{}l@{}} What was the maximum group \\ operating profit in 2019?\end{tabular} & 283.8 \\ \midrule
  \begin{tabular}[c]{@{}l@{}} \{'columns': \\ \texttt{[}'(In millions of dollars, except capital intensity)', 'Years ended December 31', '', ''\texttt{]}, \\ 'data': \texttt{[} \\ \texttt{[}'', '2019', '2018', '\%Chg'\texttt{]}, \\ \texttt{[}'Capital expenditures 1', '', '', ''\texttt{]}, \\ \texttt{[}'Wireless', '1,320', '1,086', '22'\texttt{]}, \texttt{[}'Cable', '1,153', '1,429', '(19)'\texttt{]}, \\ \texttt{[}'Media', '102', '90', '13'\texttt{]}, \\ \texttt{[}'Corporate', '232', '185', '25'\texttt{]}, \texttt{[}'Capital expenditures 1', '2,807', '2,790', '1'\texttt{]}, \\ \texttt{[}'Capital intensity 2', '18.6\%', '18.5\%', '0.1 pts'\texttt{]}\texttt{]}\} \end{tabular} &
  \begin{tabular}[c]{@{}l@{}} What was the increase \\ / (decrease) in wireless capital \\ expenditure from 2018 to 2019?\end{tabular} & 234 \\ \midrule
  \begin{tabular}[c]{@{}l@{}} \{'columns': \texttt{[}' ', 'June 30,', ' '\texttt{]},'data': \texttt{[}\texttt{[}' ', '2019', '2018'\texttt{]}, \\ \texttt{[}'Deferred tax assets', '', ''\texttt{]}, \\ \texttt{[}'Non-capital loss carryforwards', '\$161,119', '\$129,436'\texttt{]}, \\ \texttt{[}'Capital loss carryforwards', '155', '417'\texttt{]}, \\ \texttt{[}'Undeducted scientific research and development expenses', '137,253', '123,114'\texttt{]}, \\ \texttt{[}'Depreciation and amortization', '683,777', '829,369'\texttt{]}, \\ \texttt{[}'Restructuring costs and other reserves', '17,845', '17,202'\texttt{]}, \\ \texttt{[}'Deferred revenue', '53,254', '62,726'\texttt{]}, \\ \texttt{[}'Other', '59,584', '57,461'\texttt{]}, \\ \texttt{[}'Total deferred tax asset', '\$1,112,987', '\$1,219,725'\texttt{]}, \\ \texttt{[}'Valuation Allowance', '\$(77,328)', '\$(80,924)'\texttt{]}, \\ \texttt{[}'Deferred tax liabilities', '', ''\texttt{]}, \\ \texttt{[}'Scientific research and development tax credits', '\$(14,482)', '\$(13,342)'\texttt{]}, \\ \texttt{[}'Other', '(72,599)', '(82,668)'\texttt{]}, \\ \texttt{[}'Deferred tax liabilities', '\$(87,081)', '\$(96,010)'\texttt{]}, \\ \texttt{[}'Net deferred tax asset', '\$948,578', '\$1,042,791'\texttt{]}, \texttt{[}'Comprised of:', '', ''\texttt{]}, \\ \texttt{[}'Long-term assets', '1,004,450', '1,122,729'\texttt{]}, \\ \texttt{[}'Long-term liabilities', '(55,872)', '(79,938)'\texttt{]}, \\  \texttt{[}'', '\$948,578', '\$1,042,791'\texttt{]}\texttt{]}\} \end{tabular} &
  \begin{tabular}[c]{@{}l@{}} What is the total assets \\ as of June 30, 2019?\end{tabular} & 948,578 \\ \bottomrule
\end{tabular}%
}
\caption{TAT-QA Dataset examples}
\label{tab:tatqa_examples}
\end{table*}
}
% \usepackage{booktabs}
% \usepackage{graphicx}
{
\begin{table*}[t]
\centering
\resizebox{\textwidth}{!}{%
\begin{tabular}{@{}|l|l|l|l|@{}}
\toprule
\textbf{Table} &
  \multicolumn{1}{c|}{\textbf{Question}} &
  \multicolumn{1}{c|}{\textbf{Answer}} \\ \midrule
  \begin{tabular}[c]{@{}l@{}} \{'columns': \texttt{[}'season', \\ 'tropical lows', 'tropical cyclones', 'severe tropical cyclones', \\ 'strongest storm'\texttt{]}, \\ 'data': \texttt{[}\texttt{[}'1990 - 91', 10, 10, 7, 'marian'\texttt{]}, \\ \texttt{[}'1991 - 92', 11, 10, 9, 'jane - irna'\texttt{]}, \\ \texttt{[}'1992 - 93', 6, 3, 1, 'oliver'\texttt{]}, \\ \texttt{[}'1993 - 94', 12, 11, 7, 'theodore'\texttt{]}, \\ \texttt{[}'1994 - 95', 19, 9, 6, 'chloe'\texttt{]}, \\ \texttt{[}'1995 - 96', 19, 14, 9, 'olivia'\texttt{]}, \\ \texttt{[}'1996 - 97', 15, 14, 3, 'pancho'\texttt{]}, \\ \texttt{[}'1997 - 98', 10, 9, 3, 'tiffany'\texttt{]}, \\ \texttt{[}'1998 - 99', 21, 14, 9, 'gwenda'\texttt{]}, \\ \texttt{[}'1999 - 00', 13, 12, 5, 'john / paul'\texttt{]}\texttt{]}\} \end{tabular} & \begin{tabular}[c]{@{}l@{}} What is the average number \\ of tropical cyclones per season?\end{tabular} & 10.6 \\ \midrule
  \begin{tabular}[c]{@{}l@{}} \{'columns': \texttt{[}'draw', 'artist', 'song', 'points', 'place'\texttt{]}, \\ 'data': \texttt{[}\texttt{[}1, 'niamh kavanagh', 'in your eyes', 118, 1\texttt{]}, \\ \texttt{[}2, 'suzanne bushnell', 'long gone', 54, 7\texttt{]}, \\ \texttt{[}3, 'patricia roe', 'if you changed your mind', 75, 3\texttt{]}, \\ \texttt{[}4, 'róisín ní haodha', 'mo mhúirnín óg', 34, 8\texttt{]}, \\ \texttt{[}5, 'champ', '2nd time around', 79, 2\texttt{]}, \\ \texttt{[}6, 'off the record', 'hold out', 61, 6\texttt{]}, \\ \texttt{[}7, 'dav mcnamara', 'stay', 67, 4\texttt{]}, \\ \texttt{[}8, 'perfect timing', 'why aren't we talking anyway', 62, 5\texttt{]}\texttt{]}\} \end{tabular} &
  \begin{tabular}[c]{@{}l@{}} What is the difference in points \\ between the artist \\ with the highest points \\ and the average points \\ of the top 3 artists? \end{tabular} & 35.67 \\ \midrule
  \begin{tabular}[c]{@{}l@{}} \{'columns': \texttt{[}'party', \\ 'administrative panel', 'agricultural panel', \\ 'cultural and educational panel', 'industrial and commercial panel', \\ 'labour panel', 'national university of ireland', 'university of dublin', \\ 'nominated by the taoiseach', 'total'\texttt{]}, \\ 'data': \texttt{[}\texttt{[}'fianna fáil', 2, 4, 2, 3, 5, 0, 0, 9, 25\texttt{]}, \\ \texttt{[}'fine gael', 3, 4, 3, 3, 2, 1, 0, 0, 16\texttt{]}, \\ \texttt{[}'labour party', 1, 1, 0, 1, 2, 0, 0, 0, 5\texttt{]}, \\ \texttt{[}'clann na talmhan', 0, 1, 0, 0, 0, 0, 0, 0, 1\texttt{]}, \\ \texttt{[}'independent', 1, 0, 0, 1, 1, 2, 3, 1, 9\texttt{]}, \\ \texttt{[}'total', 7, 11, 5, 9, 11, 3, 3, 11, 60\texttt{]}\texttt{]}\} \end{tabular} &
  \begin{tabular}[c]{@{}l@{}} What is the total number of seats \\ held by parties that have at \\ least 2 seats in the agricultural panel, \\ and what percentage of the total seats \\ do they represent? \end{tabular} & 41, 68.33\% \\ \bottomrule
\end{tabular}%
}
\caption{TableBench Dataset examples}
\label{tab:tablebench_examples}
\end{table*}
}
% \usepackage{booktabs}
% \usepackage{graphicx}
{
\begin{table*}[t]
\centering
\resizebox{\textwidth}{!}{%
\begin{tabular}{@{}|l|l|l|l|@{}}
\toprule
\textbf{Table} &
  \multicolumn{1}{c|}{\textbf{Question}} &
  \multicolumn{1}{c|}{\textbf{Answer}} \\ \midrule
  \begin{tabular}[c]{@{}l@{}} \{'columns': \texttt{[}'player', 'average', '100s', 'matches', 'highest score', 'runs', '50s'\texttt{]}, \\'index': \texttt{[}0, 1, 2, 3, 4, 5, 6, 7\texttt{]}, \\ 'data': \texttt{[}\texttt{[}'lionel palairet', '32.04', '1', '10.0', '103', '\$575.00', '5'\texttt{]}, \\ \texttt{[}'herbie hewett', '18.98', '0', '12.0', '67', '\$398.00', '2'\texttt{]}, \\ \texttt{[}'richard palairet', '19.52', '0', '10.0', '76', '\$273.000', '1'\texttt{]}, \\ \texttt{[}'sammy woods', '18.82', '0', '11.0', '51', '\$339.0', '1'\texttt{]}, \\ \texttt{[}'vernon hill', '12.58', '0', '9.0', '32', 'N/A', '0'\texttt{]}, \\ \texttt{[}'john challen', '25.98', '0', '9.0', '-', '\$364.0', '2'\texttt{]}, \\ \texttt{[}'george nichols', '10.56', '0', '12.0', '38', '\$222.00', '0'\texttt{]}, \\ \texttt{[}'ted tyler', '10.14', '0', '12.0', '63', '\$172.0', '1'\texttt{]}\texttt{]}\} \end{tabular} & \begin{tabular}[c]{@{}l@{}} What is the total number of \\ matches played by all players in the table, \\ and what is the average number \\ of matches per player, \\ given the total number of matches? \end{tabular} & 85, 10.63 \\ \midrule
  \begin{tabular}[c]{@{}l@{}} \{'columns': \texttt{[}'', '', 'Year Ended December 31,', 'Year Ended December 31,'\texttt{]}, \\ 'index': \texttt{[}0, 1, 2, 3, 4, 5, 6, 7, 8, 9, 10, 11, 12, 13, 14, 15, 16, 17, 18, 19, 20, 21, 22, 23, 24\texttt{]}, \\ 'data': \texttt{[}\texttt{[}'', '', '2017 (a)\%', '2016 (a)\%'\texttt{]}, \\ \texttt{[}'Nonoperating income (expense):', 'Interest income', '59', '43'\texttt{]}, \\ \texttt{[}'Operating expense:', 'Depreciation and amortization', '2,213', '2,036'\texttt{]}, \\ \texttt{[}'Nonoperating income (expense):', 'Income tax expense', '923', '1,585'\texttt{]}, \\ \texttt{[}'Operating expense:', 'Other operating expenses', '5,717', '5,477'\texttt{]}, \\ \texttt{[}'Operating revenue:', 'Passenger revenue', '\$35,494', '\$34,432'\texttt{]}, \\ \texttt{[}'Operating expense:', 'Operating income', '3,781', '4,474'\texttt{]}, \\ \texttt{[}'Nonoperating income (expense):', 'Miscellaneous, net', '(104)', '(11)'\texttt{]}, \\ \texttt{[}'Nonoperating income (expense):', 'Interest expense', '(691)', '(694)'\texttt{]}, \\ \texttt{[}'Nonoperating income (expense):', 'Interest capitalized', '87', '74'\texttt{]}, \\ \texttt{[}'Operating expense:', 'Landing fees and other rent', '2,307', '2,230'\texttt{]}, \\ \texttt{[}'Operating expense:', 'Aircraft maintenance materials and outside repairs', '1,912', '1,801'\texttt{]}, \\ \texttt{[}'Nonoperating income (expense):', 'Total nonoperating expense, net', '(650)', '(588)'\texttt{]}, \\ \texttt{[}'Operating revenue:', 'Total operating revenue', '38,917', '37,654'\texttt{]}, \\ \texttt{[}'Operating expense:', 'Total operating expense', '35,136', '33,180'\texttt{]}, \\ \texttt{[}'Nonoperating income (expense):', 'Income before income taxes', '3,131', '3,886'\texttt{]}, \\ \texttt{[}'Operating expense:', \\ 'Special charges', '181', '767'\texttt{]}, \\ \texttt{[}'Operating expense:', 'Regional capacity purchase', '2,299', '2,263'\texttt{]}, \\ \texttt{[}'Operating expense:', 'Salaries and related costs', '11,269', '10,481'\texttt{]}, \\ \texttt{[}'Operating expense:', 'Aircraft fuel', '7,120', '5,987'\texttt{]}, \\ \texttt{[}'Operating expense:', 'Distribution expenses', '1,478', '1,437'\texttt{]}, \\ \texttt{[}'Operating expense:', 'Aircraft rent', '640', '700'\texttt{]}, \\ \texttt{[}'Operating revenue:', 'Cargo', '1,147', '962'\texttt{]}, \\ \texttt{[}'Nonoperating income (expense):', 'Net income', '\$2,208', '\$2,301'\texttt{]}, \\ \texttt{[}'Nonoperating income (expense):', 'Earnings per share, diluted', '\$7.27', '\$6.96'\texttt{]}\texttt{]}\} \end{tabular} &
  \begin{tabular}[c]{@{}l@{}} What is the total nonoperating \\ expense for the year \\ ended December 31, 2017, \\ and What is the percentage change \\ it the percentage change \\ in total nonoperating income \\ from 2016 to 2017? \\ Express it as a percentage.\end{tabular} & 7860.27, -0.1446 \\ \midrule
  \begin{tabular}[c]{@{}l@{}} \{'columns': \texttt{[}'(\$ in millions)', '', ''\texttt{]}, \\ 'data': \texttt{[}\texttt{[}'For the year ended December 31:', '2019', '2018'\texttt{]}, \\ \texttt{[}'Net cash provided by/(used in) continuing operations', '', ''\texttt{]}, \\ \texttt{[}'Operating activities', '$15,200', '$15,700'\texttt{]}, \\ \texttt{[}'Investing activities', '(27,800)', '(5,050)'\texttt{]}, \\ \texttt{[}'Financing activities', '9,300', '(10,800)'\texttt{]}, \\ \texttt{[}'Effect of exchange rate changes on cash, cash equivalents and restricted cash', \\ '(172)', '(510)'\texttt{]}, \\ \texttt{[}'Net change in cash, cash equivalents and restricted cash', '\$(3,390)', '\$(650)'\texttt{]}\texttt{]}, \\ 'index': \texttt{[}0, 1, 2, 3, 4, 5, 6\texttt{]}\} \end{tabular} &
  \begin{tabular}[c]{@{}l@{}} What was the increase or decrease \\ in cash from operating activities \\ from 2018 to 2019, \\ and What is the corresponding \\ percentage change \\ during the same period? \end{tabular} & -477, -3.13 \\ \bottomrule
\end{tabular}%
}
\caption{CalTab151 Dataset examples}
\label{tab:caltab151_examples}
\end{table*}
}

Table~\ref{tab:tatqa_examples}, Table~\ref{tab:tablebench_examples} and Table~\ref{tab:caltab151_examples} demonstrate the examples of each dataset in our experiment. All tabular data are stored in the format of JSON strings.

\section{Case Study}
\label{appendix:CaseStudy}

% \usepackage{booktabs}
% \usepackage{graphicx}
{
\begin{table*}[t]
\centering
\resizebox{\textwidth}{!}{%
\begin{tabular}{@{}|c|l|l|l|@{}}
\toprule
\textbf{Error Type} &
  \multicolumn{1}{c|}{\textbf{Question}} &
  \multicolumn{1}{c|}{\textbf{Error}} \\ \midrule
\begin{tabular}[c]{@{}c@{}}\textbf{Code Execution Error} \\ \textbf{(MIX-SC)} \end{tabular} &
  \begin{tabular}[c]{@{}l@{}}What is the total number of\\ Jamaicans granted British \\ citizenship from 1997 to 2008,\\ excluding years with null \\ or missing data,\\ and how many records are valid over \\ the same period?\end{tabular} &
  valueerror: could not convert \textcolor{myred}{string to float: '3,275'} \\ \midrule
\begin{tabular}[c]{@{}c@{}} \textbf{Partial Answer} \\ \textbf{(TabSQLify)}\end{tabular} &
  \begin{tabular}[c]{@{}l@{}}What is the total amount of estimated \\ interest expense across all years, \\ and how does it compare to \\ the total amount of senior unsecured notes, \\ higher or lower?\end{tabular} &
  \begin{tabular}[c]{@{}l@{}}step 1: the estimated interest expense values are:\\
  - 2011: \$273,200 - 2012: \$252,911 - 2013: \$227,796 - 2014:\\
  \$203,042 - 2015: \$182,892 - after 2015: \$365,797\\
  summing these values: \\
  273200+252911+227796+203042+182892+365797=1407638.\\
  the total estimated interest expense is \$1,407,638.\\
  step 2: calculate the total senior unsecured notes\\
  the senior unsecured notes values are:\\
  - 2011: \$87,368 - 2012: \$126,866 - 2013: \\
  \$2014 (note: this is not a numerical value, \\ so it should be treated as 0) \\ - 2014: \$101,535 - 2015: \$677\\
  after 2015: \\ \$817,115 summing these values:\\
  87368 + 126866 + 0 + 101535 + 677 + 817115 = 1133661\\
  so, the total senior unsecured notes is \$1,133,661.\\
  step 3: compare the totals\\
  now, we compare the two totals:\\- total estimated interest expense: \$1,407,638\\
  - total senior unsecured notes: \$1,133,661\\
  \$1,407,638 is greater than \$1,133,661.\\
  answer: the total estimated interest expense is \\
  \textcolor{mygreen}{higher} than the total senior unsecured notes\end{tabular} \\ \midrule
\begin{tabular}[c]{@{}c@{}}\textbf{Partial Answer,Calculational error} \\ \textbf{(Chain-of-Table)} \end{tabular} &
  \begin{tabular}[c]{@{}l@{}}What is the number of categories \\ of aircraft listed in the table? \\ and what is the total average \\ seating capacity across the aircraft models? \\ Do not calculate the unoccupied \\ parts of the aircraft.\end{tabular} &
  \begin{tabular}[c]{@{}l@{}}The number of categories of aircraft listed in the table is 13.\\The total average seating capacity across \\ these aircraft models is 2040 \\(calculated as the sum of all values in the table: \\131 + 154 + 182 + 265 + 299 + 164 + \\183 + 216 + 270 + 318 + 232 + 292 + 101 + 143 =\\ \textcolor{myred}{2040}).\end{tabular} \\ \bottomrule
\end{tabular}%
}
\caption{Case Studies for different errors in Qwen2.5-7B-Instruct; \textcolor{mygreen}{Green} is the correct answer. \textcolor{myred}{Red} is the incorrect answer of prompt-based methods, which have been used in the Table~\ref{tab:ResultsTables}.}
\label{tab:CaseStudy}
\end{table*}
}

We analyzed the error cases of the Qwen2.5-7B-Instruct on the \benchmark dataset. The detailed cases can be found in Figure~\ref{tab:CaseStudy}. 

In the case study, it is observed that the output of LLMs only addresses part of the question. For example, in CoT-based methods, not all sub-questions in complex multi-hop reasoning are answered. We propose that large models fail to recognize the sub-questions in such multi-hop scenarios, leading to incomplete answers.

In this case, MIX-SC, a PoT-based method, produces the error message: ``could not convert string to float: '3,275'.''. The error occurs due to the presence of the non-numeric character ``,'' which prevents successful conversion to a float. This issue arises because PoT-based methods, during the code generation process, focus solely on generating the code to output the final answer, while neglecting the necessary preprocessing of table contents.

From the reasoning steps of TabSQLify and Chain-of-Table, it becomes clear that while some numbers are selected correctly, the subsequent calculations remain inaccurate. This further highlights the limitation of CoT-based methods in performing precise numerical computations.

\section{Prompts in \method}
\label{appendix:tabdsr}
In this subsection, we show the prompts used in our methods, as shown in Figures~\ref{fig:QueryDecomposer}, \ref{fig:StructureTableCleaner}, and \ref{fig:PoTBasedReasoner}.

\begin{figure*}[t]
\begin{tcolorbox}[
  colback=white, % 背景颜色为黑色
  colframe=black, % 边框颜色为黑色
  colupper=gray!70, % 标题颜色为浅灰色
  fontupper=\footnotesize, % 内容字体大小
  % fonttitle=\bfseries\footnotesize, % 标题字体
  coltitle=white, % 标题字体颜色
  coltext=black, % 内容字体颜色
  boxrule=0.5mm, % 边框厚度
  title=Prompt Template of Query Decomposer
]
You are tasked with analyzing a user query to identify the number of sub-questions it contains. Your objectives are to:\\
1. \texttt{Identify Sub-Questions}: Split the query into sub-questions based solely on conjunctions (like “and”, “or”) or punctuation (such as commas). Treat each segment as a distinct sub-question boundary.\\
2. \texttt{Count Sub-Questions}: Provide the total number of sub-questions identified.\\
3. \texttt{List Sub-Questions}: List each sub-question in order as they appear in the original query. \\

\textbf{\#\# Input Format} \\

You will receive input in JSON format with the following keys: \\
- \texttt{Query}: User query string. \\

\begin{verbatim}
```json
{
  "Query": "Query String"
}
```
\end{verbatim}

\textbf{\#\# Example Input} \\

\begin{verbatim}
```json
{
  "Query": "How much money was spent on product A and how much did product B sell in total in 2015? 
           Finally, tell me the total sales of both products for the entire year."
}
```
\end{verbatim}

\textbf{\#\# Expected Output} \\

Your output should be a JSON object containing: \\
1. \texttt{subQueryCount}: The total number of sub-questions. \\
2. \texttt{subQueries}: A list of sub-questions in their original form. \\

\textbf{\#\# Example Output} \\

\begin{verbatim}
```json
{
  "subQueryCount": 3,
  "subQueries": [
    "How much money was spent on product A",
    "how much did product B sell in total in 2015",
    "tell me the total sales of product A and product B for the entire year"
  ]
}
```
\end{verbatim}

\textbf{Important Notes} \\
- \texttt{Strict JSON Format}: Ensure the output is valid JSON that can be parsed by json.loads. \\
- \texttt{No Complex Reasoning}: Do not attempt to infer meanings, just split based on conjunctions and punctuation. \\

\textbf{\#\# Input} \\

\begin{verbatim}
```json
{{Inputs}}
```
\end{verbatim}

\end{tcolorbox}
\caption{\textbf{Query Decomposer} divides a question into several sub-questions.}
\label{fig:QueryDecomposer}
\end{figure*}

\begin{figure*}[t]
\begin{tcolorbox}[
  colback=white, % 背景颜色为黑色
  colframe=black, % 边框颜色为黑色
  colupper=gray!70, % 标题颜色为浅灰色
  fontupper=\footnotesize, % 内容字体大小
  % fonttitle=\bfseries\footnotesize, % 标题字体
  coltitle=white, % 标题字体颜色
  coltext=black, % 内容字体颜色
  boxrule=0.5mm, % 边框厚度
  title=Prompt Template of Table Sanitizer,
  width=\textwidth, % 调整宽度为页面宽度
  left=1mm, % 左侧内边距
  right=1mm % 右侧内边距
]
You are tasked with cleaning and processing a JSON-formatted table while preserving the original structure as much as possible. Your main objectives are to:\\
1. Ensure column names are unique.\\
2. Clean cell data.\\
3. Maintain consistent formatting within the table.\\

\textbf{\#\# Input Structure}\\

You are provided with a JSON-formatted table containing three fields: \texttt{columns}, \texttt{data}, and \texttt{index}.\\
- \texttt{columns}: An array of strings, each representing the name of a table column.\\
- \texttt{data}: A 2D array where each nested list represents a row in the table, with each element corresponding to the cells in that row under each column.\\
- \texttt{index}: An array of integers, each representing the index of a table row.\\
\hspace*{1em}- \texttt{Note}: Rows in the \texttt{data} field are numbered starting from 0 up to the total number of rows minus one. The header row, if present, is not included in these numbers; only rows with actual data are counted.\\

\textbf{\#\# Task Requirements}\\

You need to clean and process the table based on the following requirements:\\

1. \textbf{Columns Cleaning:}\\
\hspace*{1em}- Ensure all column names are unique.\\
\hspace*{1em}- If duplicate column names exist:\\
\hspace*{2em}- Check if the first row of data contains nested column headers.\\
\hspace*{2em}- If nested headers are present, remove this row from the data. Rename columns to ensure uniqueness while preserving the original context.\\

2. \textbf{Cell Data Cleaning:}\\
\hspace*{1em}- \textbf{Numerical Columns:}\\
\hspace*{2em}- Remove extraneous symbols (e.g., \%, \$, commas, etc.) and ensure consistent numerical formatting.\\
\hspace*{2em}- Convert these values into a numerical type (e.g., float or integer) in the output JSON format.\\
\hspace*{1em}- \textbf{Non-Numerical Columns:}\\
\hspace*{2em}- Replace invalid, empty, or missing cells (e.g., N/A, null, None) with \texttt{null}.\\

3. \textbf{Row Filtering:}\\
\hspace*{1em}- Identify and exclude rows containing summary information such as: “Total”, “Sum”, “Average”, or similar statistical descriptors.\\
\hspace*{1em}- Retain all other rows to preserve the integrity of the dataset.\\

4. \textbf{Output Structure:}\\
\hspace*{1em}- Ensure that the resulting table maintains the JSON format:\\
\hspace*{2em}- All cleaned and processed columns and rows must be included.\\
\hspace*{2em}- No essential data should be lost unless explicitly instructed to remove it (e.g., summary rows).\\

\textbf{\#\# Output Structure}\\

The output should be a JSON object with the same structure as the input, containing the cleaned and processed data:\\
- \texttt{columns}: An array of strings, each representing the name of a table column.\\
- \texttt{data}: A 2D array where each nested list represents a row in the table, with each element corresponding to the cells in that row under each column.\\
\hspace*{1em}- Ensure the output maintains the strict JSON format enclosed in \texttt{json}.\\
\hspace*{1em}- All numerical data columns must be converted to appropriate numerical types.

\textbf{\#\# Input}
\begin{verbatim}
```json
{{Inputs}}
```
\end{verbatim}
\end{tcolorbox}
\caption{\textbf{Table Sanitizer} preprocesses json-formatted table.}
\label{fig:StructureTableCleaner}
\end{figure*}

\begin{figure*}[t]
\begin{tcolorbox}[
  colback=white, % 背景颜色为黑色
  colframe=black, % 边框颜色为黑色
  colupper=gray!70, % 标题颜色为浅灰色
  fontupper=\footnotesize, % 内容字体大小
  % fonttitle=\bfseries\footnotesize, % 标题字体
  coltitle=white, % 标题字体颜色
  coltext=black, % 内容字体颜色
  boxrule=0.5mm, % 边框厚度
  title=Prompt Template of PoT-based Reasoner,
  width=\textwidth, % 调整宽度为页面宽度
  left=1mm, % 左侧内边距
  right=1mm % 右侧内边距
]
\textbf{\#\# Input format}\\
You will be provided with a valid python code containing a dict of \textbf{table\_data} with the following keys.\\
- \textbf{columns}: An array of strings, each representing the name of a table column.\\
- \textbf{data}: A 2D array where each nested list represents a row in the table, with each element corresponding to the cells in that row under each column.\\
- \textbf{index}: An array of integer or string, each representing the index of a table row.\\
\hspace*{1em}- \textbf{Note}: The rows in the data are numbered starting from 0 up to the total number of rows minus one (for example, if there are 10 rows, they would be numbered from 0 to 9). The header row, if present, is not included in these numbers; only rows with actual data are counted.\\
- \textbf{Queries}: A array of string containing the user's sub-queries or request for specific information from the table.\\

Analyze the table’s structure by recognizing the relation between columns and their respective cells in data. Use these associations to identify relevant information in each cell that pertains to the Query.\\

The input format is as follows:\\
\begin{verbatim}
```python
{{InputExample}}
```
\end{verbatim}
\textbf{\#\# Task instructions}\\

You should follow these requirements below:\\

- \textbf{Analyze the Queries}:\\
\hspace*{1em}- For each sub-query, provide the following:\\
\hspace*{2em}- \textbf{Sub-query order}: Label each sub-query in order (e.g., "Sub-query 1:", "Sub-query 2:", etc.).\\
\hspace*{2em}- \textbf{Column and row indices}: Identify the relevant columns and rows in the table that are needed to answer the sub-query.\\
\hspace*{2em}- \textbf{Python code}: For each sub-query, write the corresponding Python code to extract the relevant data and compute the answer.\\
- \textbf{Code Quality}:\\
\hspace*{1em}- The Python code must be concise, easy to understand, and modular.\\
\hspace*{1em}- If necessary, add comments for clarity.\\
\hspace*{1em}- Follow best practices for code efficiency and readability.\\
- \textbf{Data Context}:\\
\hspace*{1em}- Base your analysis entirely on the provided table data. Do not use any external data or make assumptions.\\
\hspace*{1em}- If the Query is not related to the provided table data, politely refuse and provide a response explaining why.\\
- \textbf{Handling Multiple Sub-Queries}:\\
\hspace*{1em}- For multiple sub-queries, print each sub-query’s order (e.g., "Sub-query 1:", "Sub-query 2:").\\
\hspace*{1em}- For each sub-query, identify the relevant column and row indices and extract the necessary information.\\
\hspace*{1em}- For each sub-query, generate Python code to retrieve the data from the table.\\
- \textbf{Data Type Casting}:\\
\hspace*{1em}- Identify every column in the DataFrame and cast columns to appropriate data types (e.g., int, float, object) if necessary to ensure the code executes correctly.\\
- \textbf{Output Formatting}:\\
\hspace*{1em}- Provide Python code that loads the table data using the pandas library, don't response any other description.\\
\hspace*{1em}- Ensure to load the table with command table\_df=pd.DataFrame(table\_data['data'], columns=table\_data['columns']).\\
\hspace*{1em}- If the Query involves numerical calculations, perform them using DataFrame methods to get the final answers and print the final answers.\\
\hspace*{2em}- \textbf{Print the final answers}: Ensure that the final output includes the print() function to display answers. Do not print any other description information.\\
\hspace*{2em}- \textbf{Handle numerical outputs}: For any query involving calculations, format the final answer using Python’s rounding function \textbf{round()} to ensure that results are output with exactly two decimal places.\\
Replace index\_1, index\_2, etc., with the actual indices based on the identified columns and rows. If no columns or rows are identified as relevant, return an empty array for that key.\\

\textbf{\#\# User Input}\\
\begin{verbatim}
```python
{{Inputs}}
```
\end{verbatim}
\end{tcolorbox}
\caption{\textbf{PoT-based Reasoner} integrates the sub-questions and sanitized tabular data into a unified reasoning framework.}
\label{fig:PoTBasedReasoner}
\end{figure*}

\section{Other Prompts in Experiments}
\label{appendix:experimentsPrompt}

The prompts used in Table~\ref{tab:ResultsTables} and Table~\ref{tab:ablationTable} are as follows:

\paragraph{Fine-tuning LLMs} For TableGPT2~\citep{10.1145/3654979}, prompt for answer generation is Figure~\ref{fig:TableGPT2}. For TableLLM~\citep{zhang2024tablellmenablingtabulardata}, prompt for answer generation is Figure~\ref{fig:TableLLM} and~\ref{fig:TableLLMDP}. For TableLlama~\citep{zhang-etal-2024-tablellama}, prompt for answer generation is Figure~\ref{fig:TableLlama}.

\begin{figure*}[t]
\begin{tcolorbox}[
  colback=white, % 背景颜色为黑色
  colframe=black, % 边框颜色为黑色
  colupper=gray!70, % 标题颜色为浅灰色
  fontupper=\footnotesize, % 内容字体大小
  % fonttitle=\bfseries\footnotesize, % 标题字体
  coltitle=white, % 标题字体颜色
  coltext=black, % 内容字体颜色
  boxrule=0.5mm, % 边框厚度
  title=Prompt Template of TableGPT2,
  width=\textwidth, % 调整宽度为页面宽度
  left=1mm, % 左侧内边距
  right=1mm % 右侧内边距
]
Given access to several pandas dataframes, write the Python code to answer the user's question.\\
/*\\
\textquotedbl\{var\_name\}.head(5).to\_string(index=False)\textquotedbl as follows:\\
\{df\_info\}\\
*/\\

Question: \{user\_question\}
\end{tcolorbox}
\caption{\textbf{TableGPT2\textquotesingle s Prompt for Answer Generation}}
\label{fig:TableGPT2}
\end{figure*}
\begin{figure*}[t]
\begin{tcolorbox}[
  colback=white, % 背景颜色为黑色
  colframe=black, % 边框颜色为黑色
  colupper=gray!70, % 标题颜色为浅灰色
  fontupper=\footnotesize, % 内容字体大小
  % fonttitle=\bfseries\footnotesize, % 标题字体
  coltitle=white, % 标题字体颜色
  coltext=black, % 内容字体颜色
  boxrule=0.5mm, % 边框厚度
  title=Prompt Template of TableLLM,
  width=\textwidth, % 调整宽度为页面宽度
  left=1mm, % 左侧内边距
  right=1mm % 右侧内边距
]
\lbrack INST\rbrack\\
Below are the first few lines of a CSV file. You need to write a Python program to solve the provided question.\\
Header and first few lines of CSV file:\\
\{csv\_data\}\\
Question: \{question\}\lbrack /INST\rbrack
\end{tcolorbox}
\caption{\textbf{TableLLM\textquotesingle s Prompt for Code Solution (PoT)}}
\label{fig:TableLLM}
\end{figure*}
\begin{figure*}[t]
\begin{tcolorbox}[
  colback=white, % 背景颜色为黑色
  colframe=black, % 边框颜色为黑色
  colupper=gray!70, % 标题颜色为浅灰色
  fontupper=\footnotesize, % 内容字体大小
  % fonttitle=\bfseries\footnotesize, % 标题字体
  coltitle=white, % 标题字体颜色
  coltext=black, % 内容字体颜色
  boxrule=0.5mm, % 边框厚度
  title=Prompt Template of TableLLM,
  width=\textwidth, % 调整宽度为页面宽度
  left=1mm, % 左侧内边距
  right=1mm % 右侧内边距
]
\lbrack INST\rbrack\\
Offer a thorough and accurate solution that directly addresses the Question outlined in the \lbrack Question\rbrack. The answer should follow the format below:\\
\lbrack Answer Format\rbrack\\
Final Answer: AnswerName1, AnswerName2...\\
Ensure the final answer format is the last output line and can only be in the \textquotedbl Final Answer: AnswerName1, AnswerName2...\textquotedbl form, no other form. Ensure the \textquotedbl AnswerName\textquotedbl is a number or entity name, as short as possible, without any explanation.\\
\textbf{\#\#\# \lbrack Table Text\rbrack}\\
There is a table with no title.\\
\textbf{\#\#\# \lbrack Table\rbrack}\\
\textasciigrave\textasciigrave\textasciigrave \\
\{table\_in\_csv\}\\
\textasciigrave\textasciigrave\textasciigrave \\
\textbf{\#\#\# \lbrack Question\rbrack}\\
\lbrack question\rbrack\\
\textbf{\#\#\# \lbrack Solution\rbrack \lbrack INST/\rbrack}
\end{tcolorbox}
\caption{\textbf{TableLLM\textquotesingle s Prompt for Text Answer (DP)}}
\label{fig:TableLLMDP}
\end{figure*}
\begin{figure*}[t]
\begin{tcolorbox}[
  colback=white, % 背景颜色为黑色
  colframe=black, % 边框颜色为黑色
  colupper=gray!70, % 标题颜色为浅灰色
  fontupper=\footnotesize, % 内容字体大小
  % fonttitle=\bfseries\footnotesize, % 标题字体
  coltitle=white, % 标题字体颜色
  coltext=black, % 内容字体颜色
  boxrule=0.5mm, % 边框厚度
  title=Prompt Template of TableLlama,
  width=\textwidth, % 调整宽度为页面宽度
  left=1mm, % 左侧内边距
  right=1mm % 右侧内边距
]
Below is an instruction that describes a task, paired with an input that provides further context. Write a response that appropriately completes the request.\\
\textbf{\#\#\# Instruction}:\\
This is a table QA task. The goal of this task is to answer the question given the table.\\
\textbf{\#\#\# Input}:\\
\{input\}\\
\textbf{\#\#\# Question}:\\
{question}\\
\textbf{\#\#\# Response}:
\end{tcolorbox}
\caption{\textbf{TableLlama\textquotesingle s Prompt for Answer Generation}}
\label{fig:TableLlama}
\end{figure*}

\paragraph{Prompt-based LLMs} For Chain-of-Table~\citep{wang2024chainoftable}, TabSQLify~\citep{nahid-rafiei-2024-tabsqlify}, and MIX-SC~\citep{liu-etal-2024-rethinking}, their codes are publicly available.

\paragraph{DP, PoT, TCoT and SCoT} Prompts are from TableBench~\citep{wu2024tablebench}, as shown in Figure~\ref{fig:DP}, Figure~\ref{fig:PoT}, Figure~\ref{fig:CoT} and Figure~\ref{fig:SCoT}.

\begin{figure*}[t]
\begin{tcolorbox}[
  colback=white, % 背景颜色为黑色
  colframe=black, % 边框颜色为黑色
  colupper=gray!70, % 标题颜色为浅灰色
  fontupper=\footnotesize, % 内容字体大小
  % fonttitle=\bfseries\footnotesize, % 标题字体
  coltitle=white, % 标题字体颜色
  coltext=black, % 内容字体颜色
  boxrule=0.5mm, % 边框厚度
  title=Prompt Template of DP,
  width=\textwidth, % 调整宽度为页面宽度
  left=1mm, % 左侧内边距
  right=1mm % 右侧内边距
]
Your task is to answer questions based on the table content.\\
The answer should follow the format below:\\
\lbrack Answer Format\rbrack\\
Final Answer: AnswerName1, AnswerName2...\\
Ensure the final answer format is the last output line and can only be in the \textquotedbl Final Answer: AnswerName1, AnswerName2...\textquotedbl form, no other form. Ensure the \textquotedbl AnswerName\textquotedbl is a number or entity name, as short as possible, without any explanation.\\
Give the final answer to the question directly without any explanation.\\
Read the table below in JSON format:\\
\lbrack TABLE\rbrack\\
\{\{tableString\}\}\\
Let's get start!\\
Question: \{\{questionString\}\}
\end{tcolorbox}
\caption{\textbf{DP Prompt in TableBench}}
\label{fig:DP}
\end{figure*}
\begin{figure*}[t]
\begin{tcolorbox}[
  colback=white, % 背景颜色为黑色
  colframe=black, % 边框颜色为黑色
  colupper=gray!70, % 标题颜色为浅灰色
  fontupper=\footnotesize, % 内容字体大小
  % fonttitle=\bfseries\footnotesize, % 标题字体
  coltitle=white, % 标题字体颜色
  coltext=black, % 内容字体颜色
  boxrule=0.5mm, % 边框厚度
  title=Prompt Template of PoT,
  width=\textwidth, % 调整宽度为页面宽度
  left=1mm, % 左侧内边距
  right=1mm % 右侧内边距
]
You are a data analyst proficient in Python. Your task is to write executable Python code to analyze the table and then answer questions.\\
\lbrack Guidelines\rbrack\\
You should act following requirements below:\\
1. based on the question, write out your analytical approach, and then write Python code according to this approach.\\
2. The code needs to be concise and easy to understand, and if necessary, add comments for clarification.\\
3. Code blocks need to strictly start with \textasciigrave\textasciigrave\textasciigrave python and end with \textasciigrave\textasciigrave\textasciigrave .\\
4. Your analysis must be based entirely on the above data. If the user's question is not related to data analysis, please politely refuse.\\
5. You need to generate executable code. If there are results to be presented, please use the print function; if there are charts, please use the matplotlib library to draw them.\\
6. Ensure to load the table with command \textasciigrave\textasciigrave\textasciigrave \texttt{df = pd.read\_csv('table.csv')}\textasciigrave\textasciigrave\textasciigrave .\\
The answer should follow the format below:\\
\lbrack Answer Format\rbrack\\
Final Answer: AnswerName1, AnswerName2...\\
Ensure the final answer format is the last output line and can only be in the \textquotedbl Final Answer: AnswerName1, AnswerName2...\textquotedbl form, no other form. Ensure the \textquotedbl AnswerName\textquotedbl is a number or entity name, as short as possible, without any explanation.\\
Let's think step by step and then generate python code to analyze table and present the final answer to the question.\\
Read the table below in JSON format:\\
\lbrack TABLE\rbrack\\
\{\{tableString\}\}\\
Let's get start!\\
Question: \{\{questionString\}\}
\end{tcolorbox}
\caption{\textbf{PoT Prompt in TableBench}}
\label{fig:PoT}
\end{figure*}
\begin{figure*}[t]
\begin{tcolorbox}[
  colback=white, % 背景颜色为黑色
  colframe=black, % 边框颜色为黑色
  colupper=gray!70, % 标题颜色为浅灰色
  fontupper=\footnotesize, % 内容字体大小
  coltitle=white, % 标题字体颜色
  coltext=black, % 内容字体颜色
  boxrule=0.5mm, % 边框厚度
  title=Prompt Template of TCoT,
  width=\textwidth, % 调整宽度为页面宽度
  left=1mm, % 左侧内边距
  right=1mm % 右侧内边距
]
You are a table analyst. Your task is to answer questions based on the table content.\\
The answer should follow the format below:\\
\lbrack Answer Format\rbrack\\
Final Answer: AnswerName1, AnswerName2...\\
Ensure the final answer format is the last output line and can only be in the \textquotedbl Final Answer: AnswerName1, AnswerName2...\textquotedbl form, no other form. Ensure the \textquotedbl AnswerName\textquotedbl is a number or entity name, as short as possible, without any explanation.\\
Let's think step by step and then give the final answer to the question.\\
Read the table below in JSON format:\\
\lbrack TABLE\rbrack\\
\{\{tableString\}\}\\
Let's get start!\\
Question: \{\{questionString\}\}
\end{tcolorbox}
\caption{\textbf{TCoT Prompt in TableBench}}
\label{fig:CoT}
\end{figure*}
\begin{figure*}[t]
\begin{tcolorbox}[
  colback=white, % 背景颜色为黑色
  colframe=black, % 边框颜色为黑色
  colupper=gray!70, % 标题颜色为浅灰色
  fontupper=\footnotesize, % 内容字体大小
  % fonttitle=\bfseries\footnotesize, % 标题字体
  coltitle=white, % 标题字体颜色
  coltext=black, % 内容字体颜色
  boxrule=0.5mm, % 边框厚度
  title=Prompt Template of SCoT,
  width=\textwidth, % 调整宽度为页面宽度
  left=1mm, % 左侧内边距
  right=1mm % 右侧内边距
]
You are a table analyst. Your task is to utilize the Python package 'pandas' to analyze the table and then answer questions.\\
\lbrack Guidelines\rbrack\\
You should act in following patterns step by step to analyze the table and then give the final answer:\\
Patterns\rbrack\\
Thought: You should always think about what to do to interact with Python code base on Result\\
Action: the action can **ONLY** be single line python code\\
Result: Simulate the result of the execution of the python code in Action, analyse that result and decide whether to continue or not \\
(This thought/Action/Result can repeat N times) \verb|\n\n\n|The answer should follow the format below:\\
\lbrack Answer Format\rbrack\\
Final Answer: AnswerName1, AnswerName2...\\
Ensure the final answer format is the last output line and can only be in the \textquotedbl Final Answer: AnswerName1, AnswerName2...\textquotedbl form, no other form. Ensure the \textquotedbl AnswerName\textquotedbl is a number or entity name, as short as possible, without any explanation.\\
Let's think step by step and then give the final answer to the question. \\
Ensure to have a concluding thought that verifies the table, observations and the question before giving the final answer. \\
Read the table below in JSON format:\\
\lbrack TABLE\rbrack\\
\{tableString\}\\
Let's get start!\\
Question: \{questionString\}
\end{tcolorbox}
\caption{\textbf{SCoT Prompt in TableBench}}
\label{fig:SCoT}
\end{figure*}

\section{Appendix: Failure Analysis}

To better understand the robustness and limitations of our TabDSR system, we conducted a comprehensive failure analysis covering all three agents: Decomposer (D), Sanitizer (S), and Executor (R). For each module, a failure leads to a fallback behavior, potentially degrading the final performance. The table~\ref{tab8:compolent} below reports the failure rates for each module (S in Table~\ref{tab8:s}, R in Table~\ref{tab8:r}) across three datasets:

\begin{table*}[h]
\centering
\begin{tabular}{lccc}
\toprule
\textbf{Component} & \textbf{TatQa} & \textbf{TableBench} & \textbf{CalTab151} \\
\midrule
Decomposer (D) & 0.00 & 0.00 & 0.00 \\
Sanitizer (S)  & 0.03 & 0.03 & 0.01 \\
Executor (R)   & 0.15 & 0.15 & 0.17 \\
\bottomrule
\end{tabular}
\caption{Failure rates of each component in TabDSR.}
\label{tab8:compolent}
\end{table*}

\subsection*{Sanitizer (S) Error Types}
The Sanitizer component encountered parsing issues when converting JSON outputs into structured tables. The most common errors were \texttt{JSONDecodeError} and \texttt{ValueError}.

\begin{table*}[h]
\centering
\begin{tabular}{lccc}
\toprule
\textbf{Error Type} & \textbf{TatQa} & \textbf{TableBench} & \textbf{CalTab151} \\
\midrule
JSONDecodeError   & 14 & 3  & 1  \\
ValueError        & 10 & 9  & 0  \\
AttributeError    & 0  & 1  & 0  \\
\bottomrule
\end{tabular}
\caption{Sanitizer (S) error types across datasets.}
\label{tab8:s}
\end{table*}

\subsection*{Executor (R) Error Types}
The Executor agent, responsible for code execution, had the highest failure rate. The frequent exceptions—such as \texttt{ValueError}, \texttt{KeyError}, and \texttt{TypeError}—indicate the challenges in generating robust code for table reasoning tasks.

\begin{table*}[h]
\centering
\begin{tabular}{lccc}
\toprule
\textbf{Error Type} & \textbf{TatQa} & \textbf{TableBench} & \textbf{CalTab151} \\
\midrule
ValueError         & 39 & 17 & 1  \\
KeyError           & 43 & 13 & 6  \\
SyntaxError        & 0  & 1  & 2  \\
NameError          & 3  & 10 & 5  \\
TypeError          & 13 & 23 & 8  \\
IndexError         & 7  & 8  & 3  \\
AttributeError     & 3  & 0  & 0  \\
UFuncTypeError     & 0  & 1  & 0  \\
\midrule
\textbf{Total Errors} & \textbf{108} & \textbf{74} & \textbf{25} \\
\textbf{Error Rate} (Errors / Dataset Size) & \textbf{0.15} & \textbf{0.15} & \textbf{0.17} \\
\bottomrule
\end{tabular}
\caption{Executor (R) error types and frequencies.}
\label{tab8:r}
\end{table*}

\subsection*{Discussion}

Our analysis reveals that the \textbf{Decomposer (D)} component was highly robust, with zero failures across all datasets, validating the structural soundness of our question decomposition pipeline.

The \textbf{Sanitizer (S)} exhibited low but non-negligible failure rates (1\%--3\%), mostly due to minor formatting or schema inconsistencies in the JSON output. These could potentially be mitigated by enhancing schema conformity or introducing lightweight JSON repair strategies.

The \textbf{Executor (R)} was the most failure-prone module, with 15\%--17\% of samples triggering code execution errors. This reflects the inherent difficulty of generating correct Python code under diverse and noisy table inputs. Addressing these issues remains an open challenge, potentially benefiting from more constrained decoding strategies or runtime feedback loops.

Overall, the failure analysis offers concrete insights for future improvement of each agent in the TabDSR pipeline.

\end{document}